\pdfoutput=1

\documentclass[11pt]{article}

\usepackage[final]{acl}

\usepackage{times}
\usepackage{latexsym}

\usepackage[T1]{fontenc}

\usepackage[utf8]{inputenc}

\usepackage{microtype}

\usepackage{inconsolata}

\usepackage{graphicx}
\usepackage{xspace}
\newcommand{\eg}{e.g.\xspace}

\usepackage{subcaption}
\usepackage[utf8]{inputenc}
\usepackage[most]{tcolorbox}
\usepackage{xcolor}
\usepackage{booktabs}
\usepackage{multirow}
\usepackage{siunitx}
\usepackage{amsmath}
\usepackage{amsfonts}
\usepackage{float}

\makeatletter
\DeclareRobustCommand\onedot{\futurelet\@let@token\@onedot}
\def\@onedot{\ifx\@let@token.\else.\null\fi\xspace}

\def\eg{\emph{e.g}\onedot} 
\def\ie{\emph{i.e}\onedot}

\makeatother

\title{Group-Aware Reinforcement Learning for Output Diversity in Large Language Models}

\author{
  \textbf{Oron Anschel} \thanks{Corresponding author: oronans@amazon.com},
  \textbf{Alon Shoshan},
  \textbf{Adam Botach},
  \textbf{Shunit Haviv Hakimi},
\\
  \textbf{Asaf Gendler},
  \textbf{Emanuel Ben Baruch},
  \textbf{Nadav Bhonker},
  \textbf{Igor Kviatkovsky},
\\
  \textbf{Manoj Aggarwal},
  \textbf{Gerard Medioni}
\\
\\
  Amazon
\\
  \small{
    \{oronans, alonshos, kabotach, havivs, gendlasa, emanbb, nadavb, kviat, manojagg, medioni\}@amazon.com
  }
}

\begin{document}
\maketitle

\begin{abstract}
Large Language Models (LLMs) often suffer from mode collapse, repeatedly generating the same few completions even when many valid answers exist, limiting their diversity across a wide range of tasks. 
We introduce \textbf{Group-Aware Policy Optimization (GAPO)}, a simple extension of the recent and popular Group Relative Policy Optimization (GRPO) that computes rewards over the group as a whole. GAPO enables learning from the group-level properties such as diversity and coverage.
We demonstrate GAPO using a frequency-aware reward function that encourages uniform sampling over valid LLM completions, and show that GAPO-trained models produce valid and more diverse model responses.
Beyond this setup, GAPO generalizes to open-ended prompts and improves response diversity without compromising accuracy on standard LLM benchmarks (GSM8K, MATH, HumanEval, MMLU-Pro). Our code will be made publicly available.

\end{abstract}

\begin{figure}[t]
    \centering
    \includegraphics[width=0.47\textwidth]{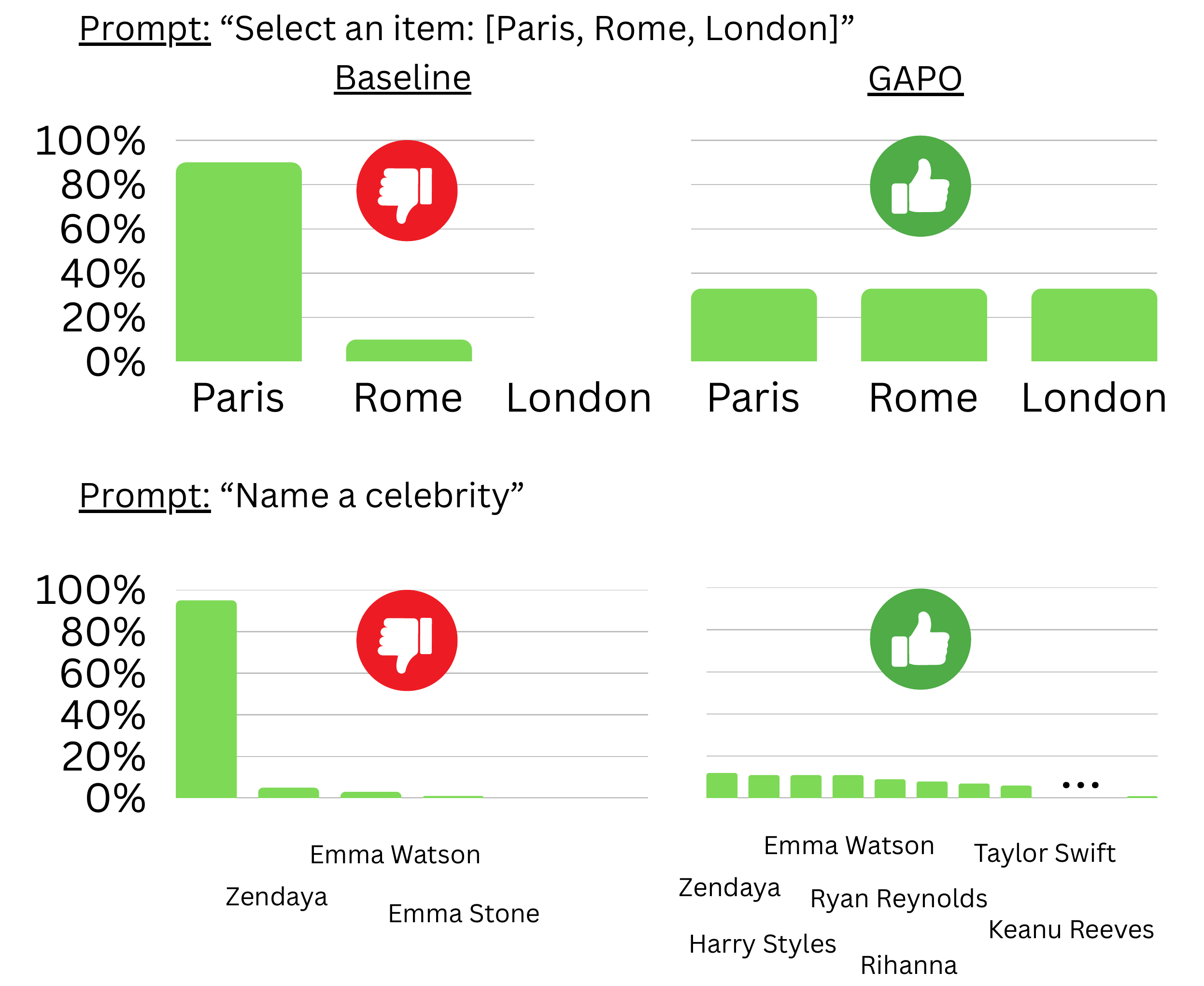}
    \caption{GAPO with a frequency-aware reward promotes output uniformity and diversity. GAPO mitigates over-representation of frequent completions and encourages more balanced generation, both in list selection tasks and open-ended prompts.}
    \label{fig:example}
\end{figure}

\section{Introduction}
Large Language Models (LLMs), particularly sophisticated instruction-following systems such as ChatGPT, Claude, Gemini, Qwen, and DeepSeek, are experiencing rapidly increasing deployment across a diverse range of real-world applications and use cases \cite{ouyang2022training,team2023gemini, yang2024qwen2,guo2025deepseek}. While reinforcement learning from human feedback (RLHF) \cite{ouyang2022training} improves factuality and alignment, it often reduces output diversity \cite{kirk2023understanding}. This limitation is especially problematic in creative or open-ended tasks, where multiple distinct completions may be equally valid.

This concerning reduction in output diversity is frequently characterized and analyzed in the literature as mode collapse \citep{o2024attributing}, a phenomenon where models exhibit a strong tendency to repeatedly generate the same limited set of responses across multiple interactions. For example, when prompted with the seemingly open-ended request \textit{tell me a joke''}, popular models like ChatGPT-4o and Claude Sonnet 3.5 frequently respond with virtually identical outputs such as: \textit{Why don't scientists trust atoms? Because they make up everything!''} \cite{jentzsch2023chatgpt}. While these responses remain both well-aligned with human preferences and linguistically fluent, such persistent repetition clearly demonstrates a problematic overconcentration of probability mass on an extremely limited subset of the vast space of possible completions, thereby significantly limiting the model's overall expressiveness and creative potential.


Recent work has identified this behavior as a consequence of the training pipeline. Both supervised fine-tuning (SFT) and RLHF have been shown to push models toward high-probability completions, leading to repeated outputs even when many valid alternatives exist~\cite{o2024attributing, kirk2023understanding}. 
While decoding strategies such as temperature scaling~\cite{ackley1985learning}, top-k sampling~\cite{fan2018hierarchical}, or nucleus (top-p) sampling~\cite{holtzman2019curious} can partially mitigate this effect, they do not address the underlying issue in the model’s probability distribution.

In this work, we take a direct approach to improving output diversity by modifying the model’s training objective rather than its decoding strategy. Specifically, we fine-tune a fully trained instruction model using LoRA~\cite{hu2022lora} with a reward function that promotes balanced sampling across valid outputs. 

To implement this, we build on Group Relative Policy Optimization (GRPO)~\cite{shao2024deepseekmath}, a reinforcement learning method that compares completions within a group to compute relative advantages. 
While GRPO assigns fixed, per-sample rewards, we extend the framework by computing rewards at the group level, allowing the model to learn distributional properties such as uniform coverage over valid outputs. We refer to this extension as \textit{Group-Aware Policy Optimization} (GAPO).

We begin by evaluating GAPO on a clean and insightful task: sampling a single item from a list of equally valid options \cite{eicher2024reducing}. Existing LLMs exhibit strong selection biases in this setting, while GAPO-trained models learn to sample nearly uniformly (Figure~\ref{fig:example}).
On open-ended prompts such as \textit{``name a city''}, \textit{``suggest a food''}, or \textit{``name a celebrity''}, GAPO generates significantly more diverse responses, even in categories unseen during training.

Finally, we demonstrate that GAPO enhances diversity in creative writing tasks such as poetry, storytelling, and dialogue while preserving coherence, as measured by accuracy on the GSM8K, MATH, HumanEval and MMLU-Pro datasets.

\paragraph{Our contributions are as follows:}
\begin{itemize}
    \item We introduce \textbf{Group-Aware Policy Optimization (GAPO)}, an extension of GRPO that defines rewards over a group of completions, enabling learning from group-level properties such as output diversity and coverage.
    
    \item We design a \textbf{frequency-aware reward function} that encourages uniform sampling over valid completions, directly addressing mode collapse without changing the model architecture or decoding strategy.

    \item We show that GAPO-trained models achieve near-uniform sampling when prompted to select items from lists, and generate substantially more diverse outputs in open-ended prompts.
    
    \item We demonstrate that GAPO \textbf{improves diversity in creative writing} tasks while maintaining coherence, as validated by performance on standard benchmarks.
\end{itemize}

\begin{figure*}[htb]
\setlength{\parindent}{0pt}
\setlength{\tabcolsep}{0.5pt}
\footnotesize
\begin{tabular}{l c r}
\begin{minipage}[t]{0.245\textwidth}
  \begin{tcolorbox}[
      colback=black!5,
      colframe=black,
      arc=2pt,
      boxrule=1pt,
      left=4pt, right=4pt, top=4pt, bottom=4pt,
      title=\textbf{Instruction Variant 1},
      fonttitle={\bfseries\color{white}},
      colbacktitle=black,
      coltitle=white,
    ]
    "Please select one of the items [Canada, ..., Spain]"
  \end{tcolorbox}
\end{minipage}
&
\begin{minipage}[t]{0.245\textwidth}
  \begin{tcolorbox}[
      colback=black!5,
      colframe=black,
      arc=2pt,
      boxrule=1pt,
      left=4pt, right=4pt, top=4pt, bottom=4pt,
      title=\textbf{Instruction Variant 2},
      fonttitle={\bfseries\color{white}},
      colbacktitle=black,
      coltitle=white,
    ]
    "Sample one item out of [Canada, ..., Spain]"
  \end{tcolorbox}
\end{minipage}
&
\begin{minipage}[t]{0.49\textwidth}
  \begin{tcolorbox}[
      colback=black!5,
      colframe=black,
      arc=2pt,
      boxrule=1pt,
      left=4pt, right=4pt, top=4pt, bottom=4pt,
      title=\textbf{Instruction Variant 1 - Shuffled List},
      fonttitle={\bfseries\color{white}},
      colbacktitle=black,
      coltitle=white,
    ]
    "Please select one of the following items: \textbf{shuffle([Canada, Mexico, ..., Spain])}"
  \end{tcolorbox}
\end{minipage}
\end{tabular}

\begin{subfigure}[b]{0.49\textwidth}
  \includegraphics[width=0.495\textwidth]{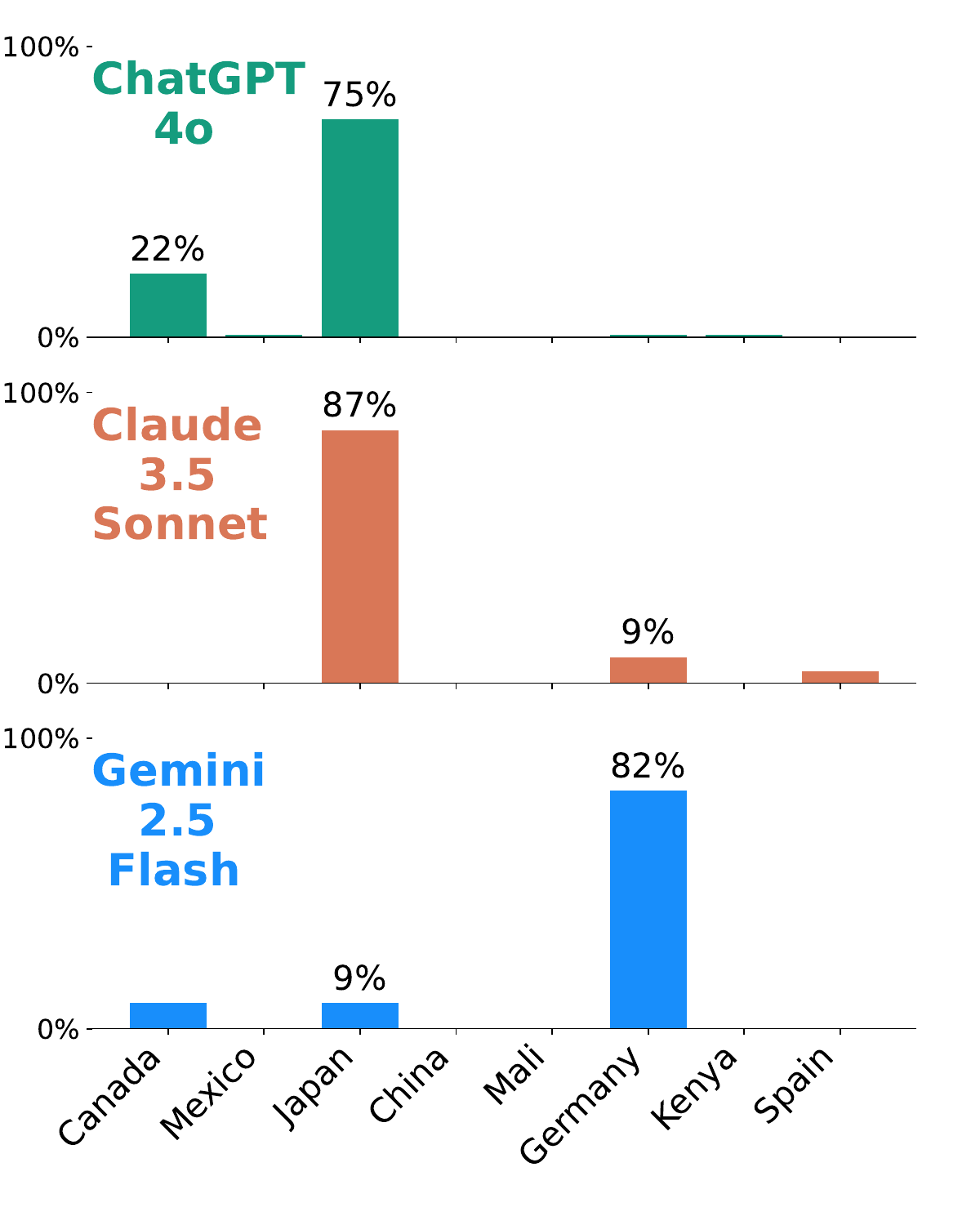}
  \includegraphics[width=0.495\textwidth]{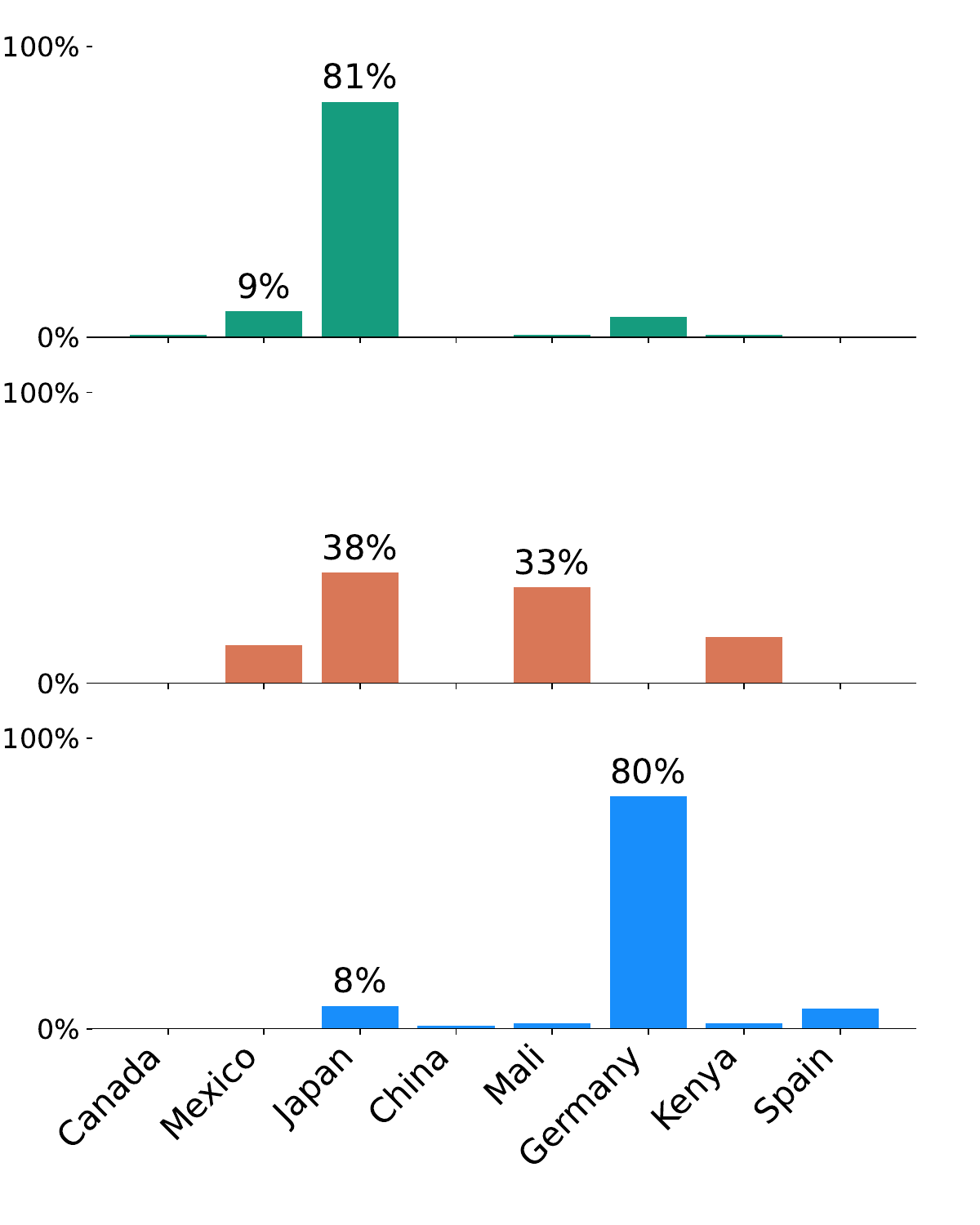}
  \footnotesize
  \caption{Same list - different instructions.}
  \label{fig:bias_no_shuffle}
\end{subfigure}
\hfill
\begin{subfigure}[b]{0.49\textwidth}
  \includegraphics[width=0.495\textwidth]{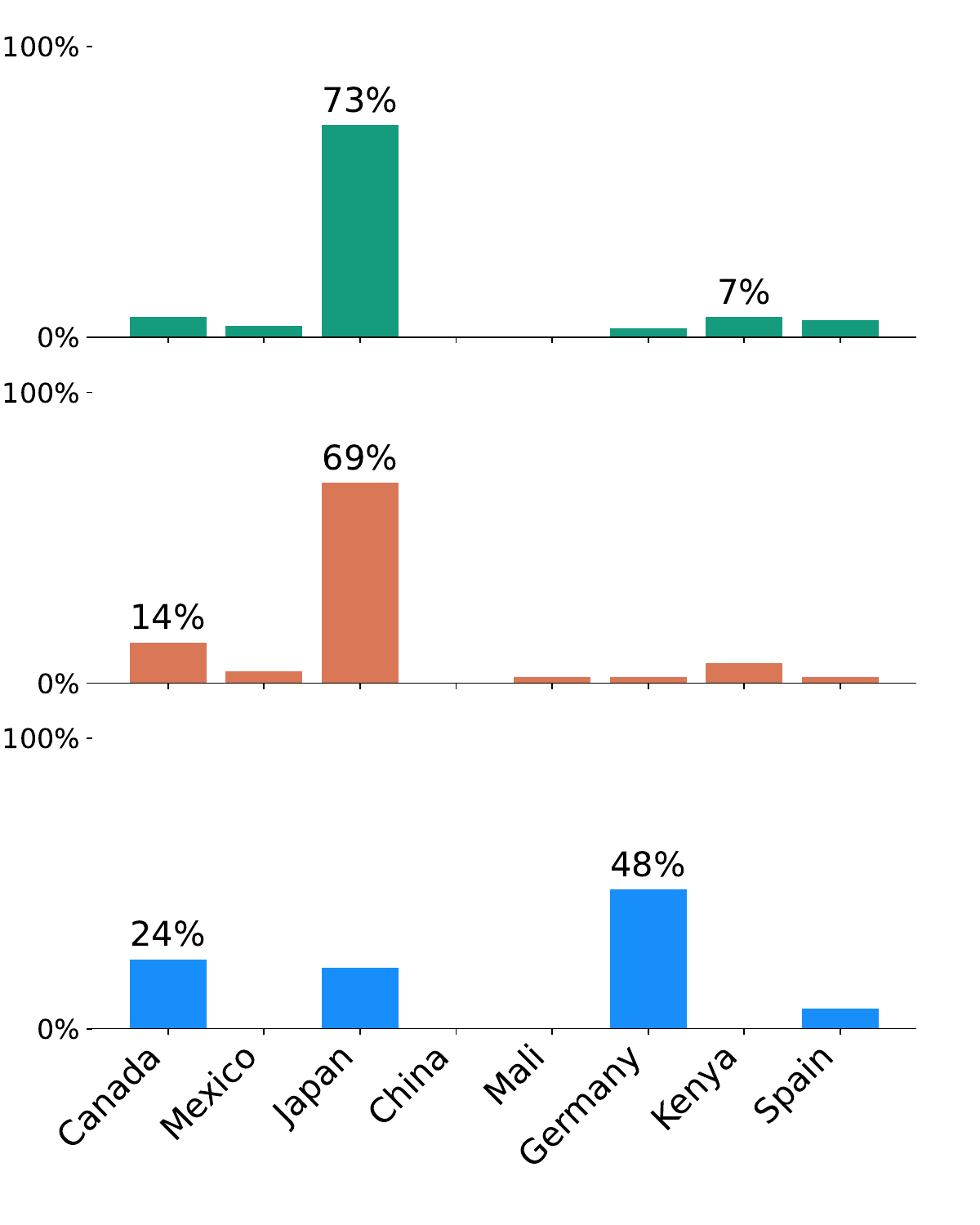}
  \includegraphics[width=0.495\textwidth]{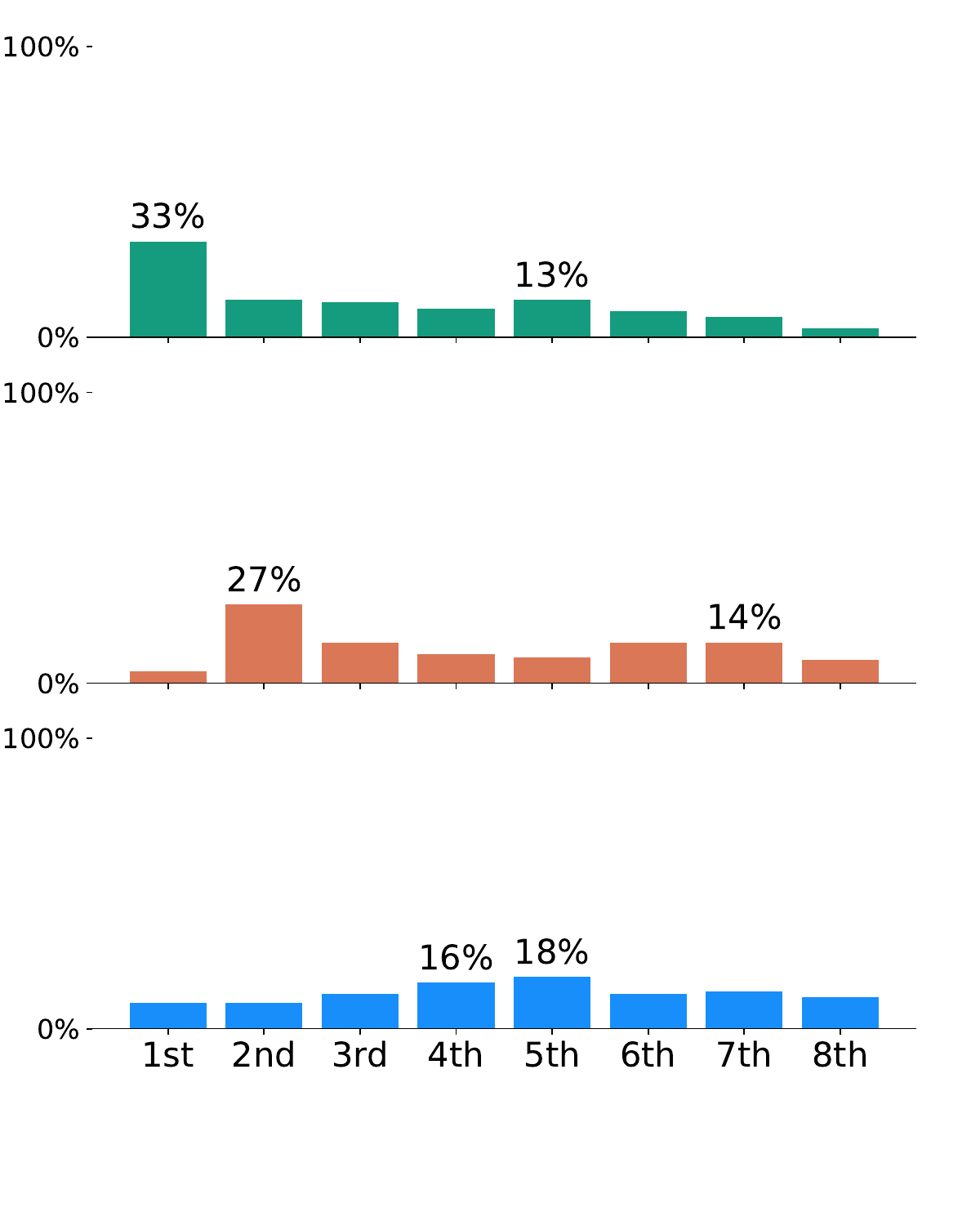}
  \footnotesize
  \caption{Same instruction - shuffled list.}
  \label{fig:bias_shuffle}
\end{subfigure}

\caption{\textbf{Bias Analysis.} (a) Using the same list of countries, we prompt each model 100 times with each of the two instruction prompts. (b) We prompt each model 100 times with the same instruction prompt, each time the list of countries is randomly shuffled. For each model the largest and second-largest probabilities are shown.}
\label{fig:bias_motivation}
\end{figure*}

\section{Motivation - Case Study}

To investigate distributional biases in current large language models (LLMs), we evaluated ChatGPT 4o, Claude Sonnet 3.5, and Gemini 2.5 by repeatedly prompting them with prompts such as: \textit{``Sample one item out of [Canada, Mexico, ..., Spain]''} and recording the output distribution.

In Figure~\ref{fig:bias_no_shuffle}, under Instruction Variant 1, \textit{``Please select one of the items''}, ChatGPT 4o and Claude favor “Japan” (75\%, 87\%), while Gemini prefers “Germany” (82\%). 
Switching to Instruction Variant 2, \textit{``Sample one item out of...''}, we observe that Claude changes their dominant choice, suggesting the presence of contextual bias.

In Figure~\ref{fig:bias_shuffle}, we probe positional bias by shuffling the list while keeping the instruction fixed. 
ChatGPT 4o continues to favor “Japan” (73\%) regardless of position, suggesting item-specific bias. 
Claude still prefers “Japan” (69\%), while Gemini's bias towards “Germany” diminishes (48\%).
Index-wise distributions show that ChatGPT favors the first item (33\%), Claude avoids it, and Gemini prefers mid-list positions, demonstrating a level of positional bias.
These results illustrate that modern LLMs exhibit item-specific, positional, and contextual biases, and often collapse onto a small subset of valid responses.

Motivated by these findings, we developed GAPO, a reinforcement learning method that, when coupled with a frequency-aware reward, encourages balanced sampling across valid completions. 
As shown in our experiments, GAPO mitigates these biases, thereby improving diversity in both structured tasks and open-ended generation.

\section{Related Work}

Neural text degeneration, where models produce repetitive and low-diversity outputs, was first identified by \citet{holtzman2019curious}. To address it, researchers proposed stochastic decoding methods like top-k, top-p, and min-p sampling \cite{fan2018hierarchical, holtzman2019curious, nguyen2024turning}, often paired with temperature scaling \cite{ackley1985learning} to balance diversity and precision. However, these are inference time fixes that do not alter the model's distribution.

Reinforcement learning is widely used to align language models with human preferences \cite{ziegler2019fine}, but often reduces output diversity \cite{kirk2023understanding}. To counter this, Welleck et al.\ \cite{welleck2019neural} proposed unlikelihood training, while Bowman et al.\ \cite{bowman2015generating} used mutual information to encourage diversity.

Entropy regularization has recently gained attention for improving diversity in supervised and RL settings. Approaches include entropy-regularized RL \cite{tiapkin2024generative}, diversity-aware DPO variants \cite{rafailov2023direct, slocumdiverse}, entropy-regularized fine-tuning \cite{li2024entropic}, and GDPO \cite{kwon2024gdpo}, which uses generative flow networks to promote diversity. In contrast, our approach promotes diversity by directly encouraging uniform probability over correct answers.

\begin{figure*}[htbp]
    \centering
    \includegraphics[width=1\textwidth]{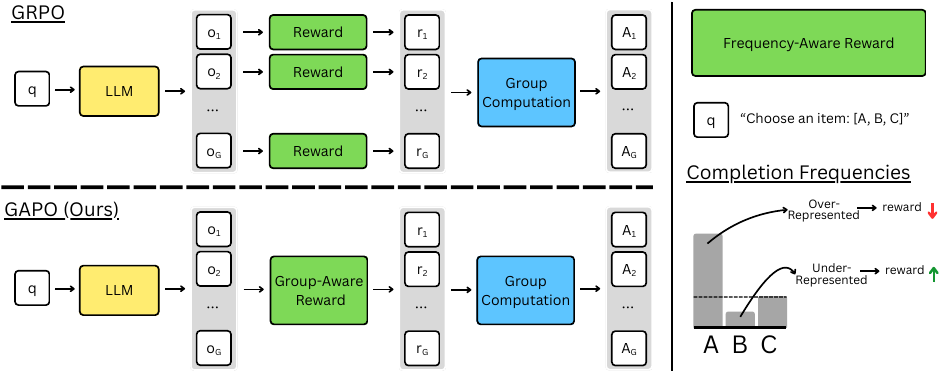}
    \caption{\textbf{GRPO vs. GAPO}. Comparison between GRPO and GAPO (left) and illustration of frequency-aware rewards (right). GRPO assigns per-sample rewards, while GAPO computes rewards over the whole group, enabling distributional signals such as diversity and coverage. Our frequency-aware reward function penalizes overrepresented outputs and boosts underrepresented ones, guiding the model toward uniform sampling over equally valid responses.
}
    \label{fig:gapo_vs_grpo}
\end{figure*}

\section{Preliminaries}
\paragraph{Group Relative Policy Optimization}
In \citet{shao2024deepseekmath} the authors presented the Group Relative Policy Optimization (GRPO) framework for optimizing language models for math and coding challenges. GRPO optimizes 
LLMs policies by estimating advantages in a group-relative manner, without relying on a value function. Below, we summarize the aspects of GRPO relevant to our work, following \citet{shao2024deepseekmath} notations. 



\paragraph{Sampling and Rollouts}  
Optimization begins by sampling a query \( q \) from the data distribution \( P(Q) \), and generating a group of \( G \) rollouts \( \{o_i\}_{i=1}^G \) using the old policy \( \pi_{\theta_{\text{old}}} \) as in \cite{schulman2015trust}. Each rollout \( o_i = (o_{i,1}, \dots, o_{i,|o_i|}) \) is a sequence of tokens generated autoregressively, where \( |o_i| \) denotes its length.

\paragraph{Rewards}  
For each rollout, we compute a scalar reward \( r_i = R(o_i) \) using a reward function \( R \), typically defined per rollout. Let \( \mathbf{r} = (r_1, \dots, r_G) \) denote the group reward vector, with mean \( \bar{r} \) and standard deviation \( \sigma_r \).


\paragraph{Advantage Estimation}  
Outcome supervision assigns each rollout an advantage based on its normalized reward within the group:
\[
\hat{A}_{i,t} = \frac{r_i - \bar{r}}{\sigma_r} 
\quad\text{for } t = 1, \dots, |o_i|.
\]

\paragraph{Policy Update}  
We then compute per-token importance sampling ratios:
\[
\rho_{i,t}(\theta) = \frac{\pi_{\theta}(o_{i,t} \mid q, o_{i,<t})}
{\pi_{\theta_{\text{old}}}(o_{i,t} \mid q, o_{i,<t})},
\]
where \( o_{i,<t} = (o_{i,1}, \dots, o_{i,t-1}) \) denotes the token prefix. Following \citet{schulman2017proximal}, we clip these ratios using a hyperparameter \( \epsilon > 0 \) to constrain the update magnitude:
\[
\begin{aligned}
\mathcal{L}_{\text{clip}}(\theta)
&= \frac{1}{G} \sum_{i=1}^G \frac{1}{|o_i|} \sum_{t=1}^{|o_i|} \min \Big\{
\rho_{i,t}(\theta) \hat{A}_{i,t}, \\
&\quad \text{clip}\big(\rho_{i,t}(\theta), 1 - \epsilon, 1 + \epsilon\big) \hat{A}_{i,t} \Big\}.
\end{aligned}
\]

\paragraph{Objective}  
The final GRPO objective combines the clipped surrogate with a KL penalty:
\[
\mathcal{J}_{\text{GRPO}}(\theta)
= \underbrace{\mathcal{L}_{\text{clip}}(\theta)}_{\text{clipped surrogate}}
- \beta \underbrace{D_{\text{KL}}[\pi_\theta \,\|\, \pi_{\text{ref}}]}_{\text{KL penalty}}.
\]
Here, \( \beta > 0 \) controls the trade-off between policy improvement and divergence from the fixed reference policy \( \pi_{\text{ref}} \). The KL term \( D_{\text{KL}}[\pi_\theta \,\|\, \pi_{\text{ref}}] \) measures the average KL divergence between the current and reference policies over the rollout distribution.

\section{Group Aware Policy Optimization}
\label{sec:gapo}

\paragraph{Group-Aware Policy Optimization (GAPO)}  
GAPO introduces a simple yet effective modification to the GRPO framework: the reward is computed jointly across the group of rollouts rather than independently per rollout (Figure~\ref{fig:gapo_vs_grpo}). This change allows the reward function to capture group-level properties—such as diversity or sampling balance—without altering the policy architecture, optimization objective, or training procedure.

Formally, for a group of rollouts \( \mathbf{o} = \{o_1,\dots,o_G\} \), the reward assigned to rollout \( i \) is
\[
r_i = \tilde{R}(\mathbf{o})_i,
\]
where \( \tilde{R}(\mathbf{o}) \in \mathbb{R}^{G} \) is a vector of group-aware rewards computed over the full set.

\begin{figure*}[h]
  \centering
  \footnotesize
  \begin{tcolorbox}[
        width=0.95\textwidth,
        colback=black!5,
        colframe=black,
        arc=2pt,
        boxrule=1pt,
        left=4pt, right=4pt, top=4pt, bottom=4pt,
        title=\textbf{Instruction},
        fonttitle={\bfseries\color{white}},
        colbacktitle=black,
        coltitle=white
        ]
      Please select one of the following items \{list\}.\\
      Format your response as follows: \texttt{<answer>selected\_item</answer>}.
  \end{tcolorbox}
    \begin{subfigure}[b]{0.32\textwidth}
    \includegraphics[width=\linewidth]{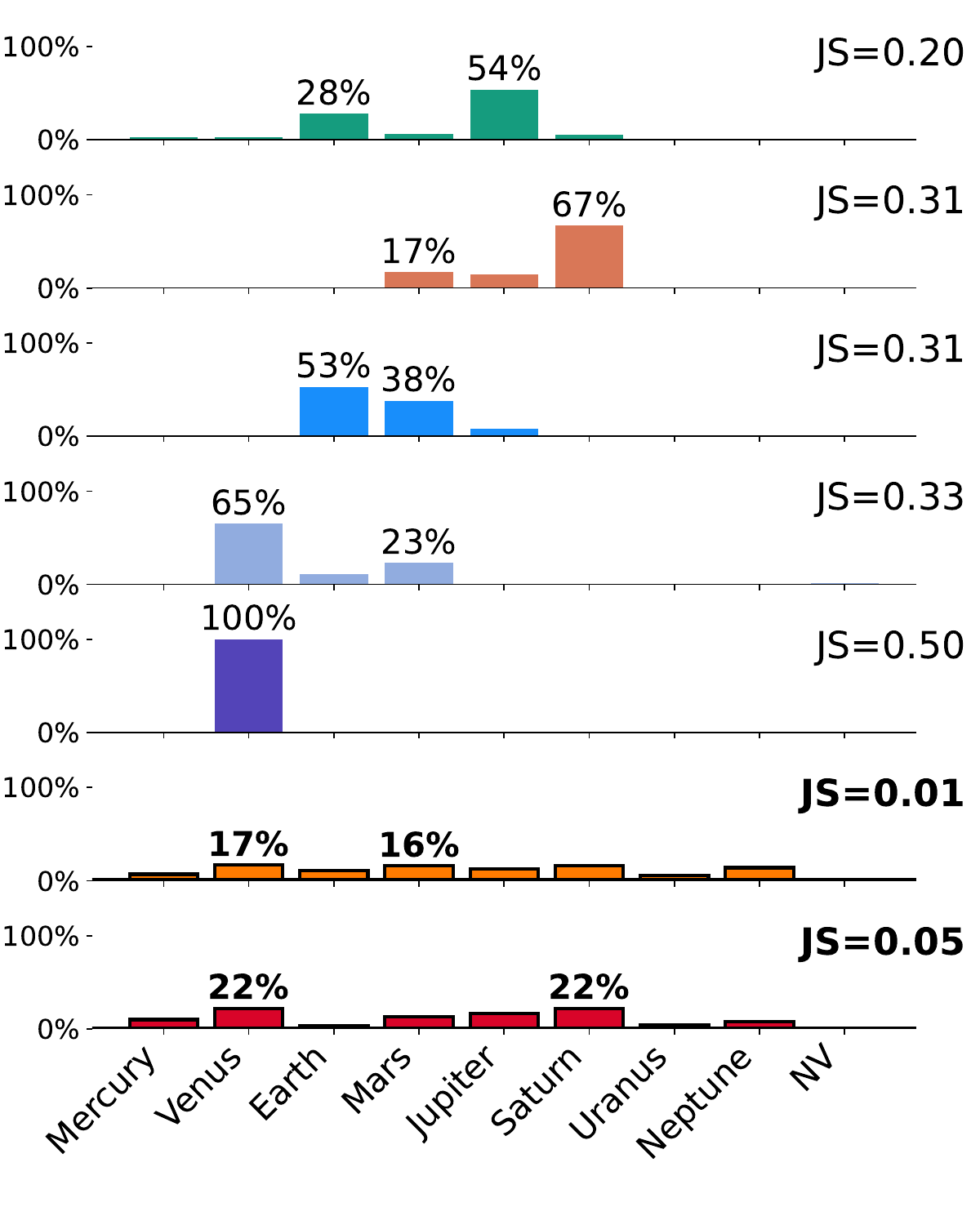}
    \caption{Planets.}
    \label{fig:planets_close_set}
  \end{subfigure}\hfill
  \begin{subfigure}[b]{0.32\textwidth}
    \includegraphics[width=\linewidth]{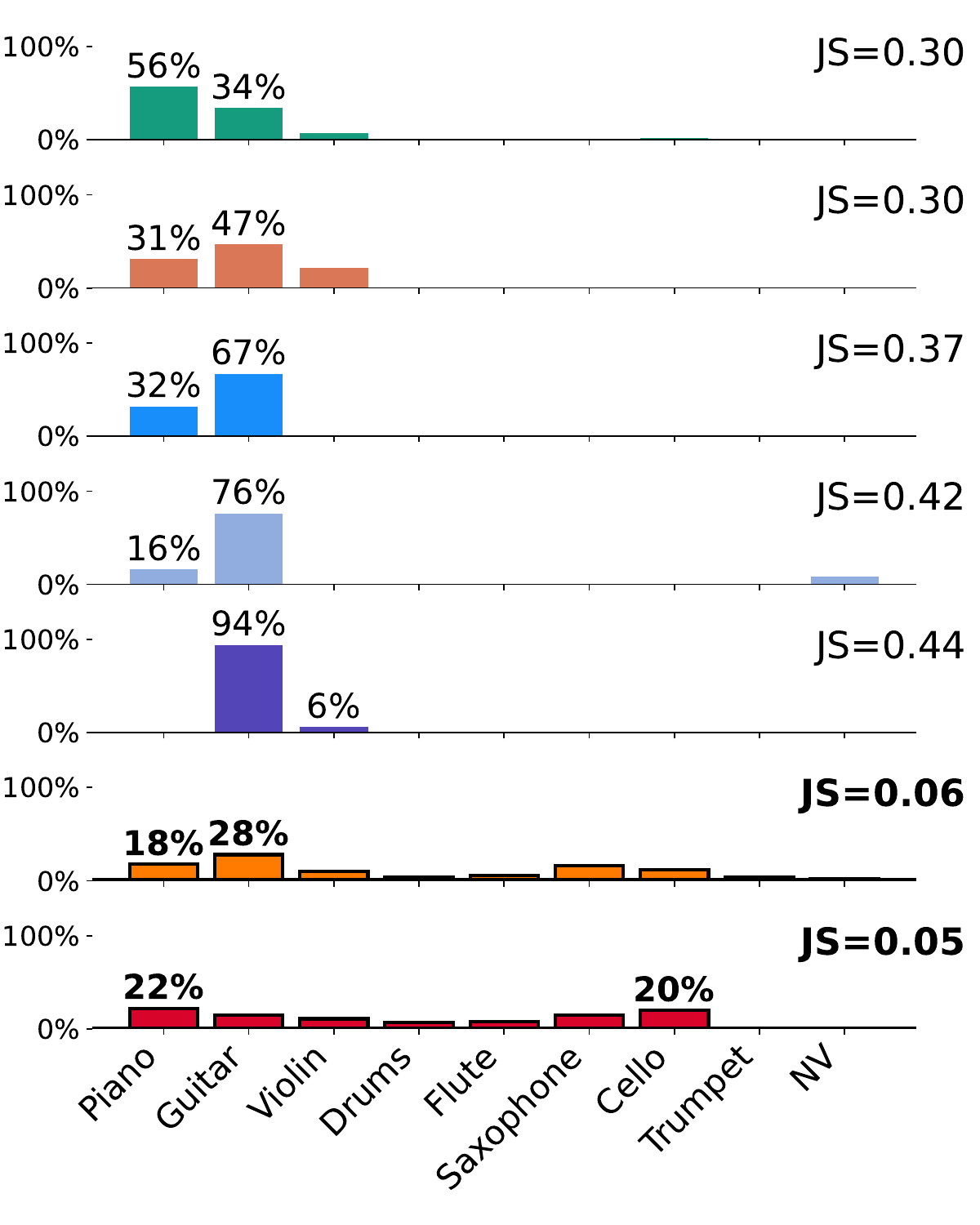}
    \caption{Music instruments.}
    \label{fig:music_close_set}
  \end{subfigure}\hfill
  \begin{subfigure}[b]{0.32\textwidth}
    \includegraphics[width=\linewidth]{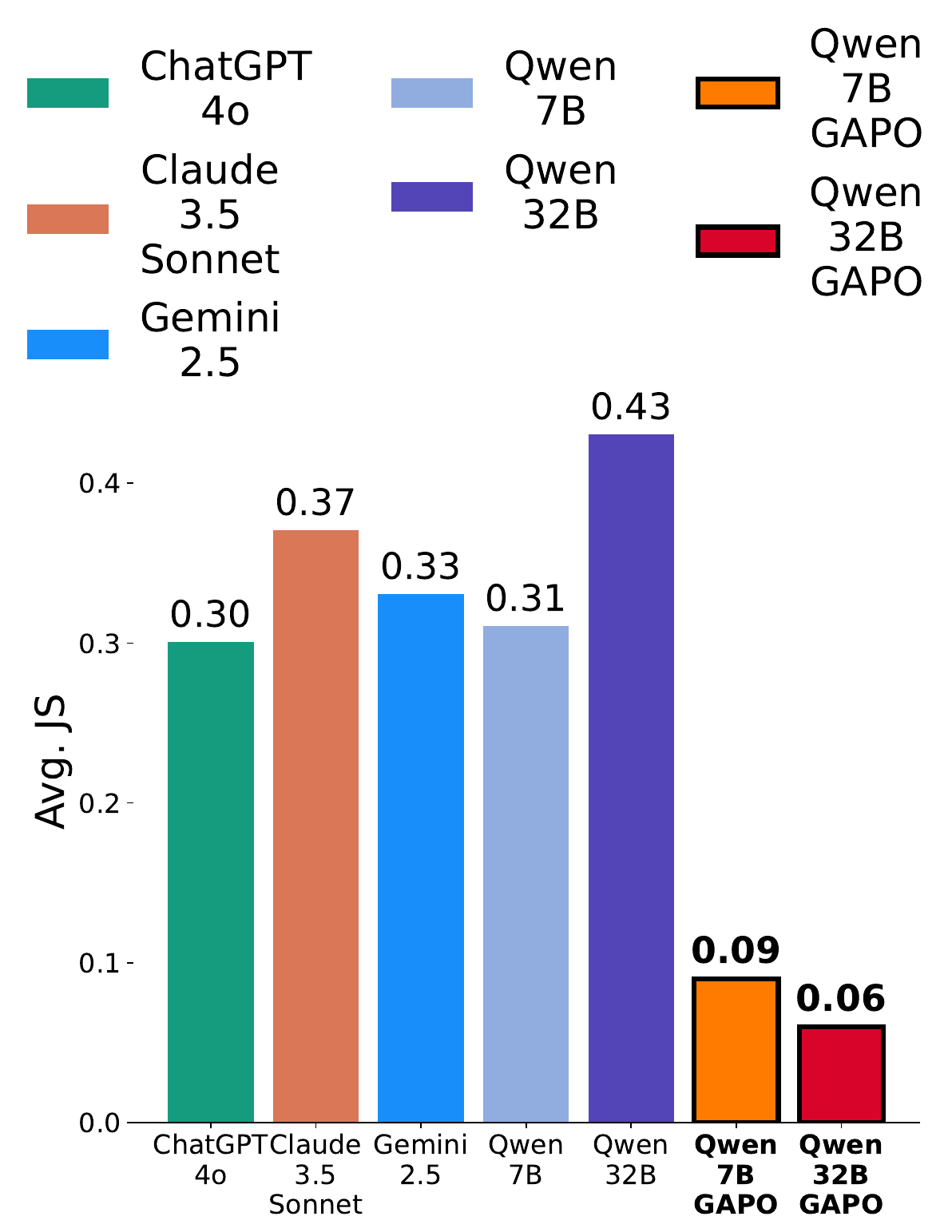}
    \caption{Average JS ($\downarrow$).}
    \label{fig:avg_js_close_set}
  \end{subfigure}
        \caption{\textbf{Sampling from a list.} Output distributions for different models when prompted to choose from a list of planets (a) or musical instruments (b). \textit{NV} denotes an invalid response not in the list. Panel~(c) shows the average Jensen-Shannon divergence (JS) from  uniform distribution across all topics.}

  \label{fig:list_uniformity}
  
\end{figure*}

\subsection{Theoretical Foundation}
\label{sec:theory}

A reward is compatible with GAPO if three standard policy-gradient conditions hold.  
(i) \emph{Parameter independence}: the reward may depend on the sampled roll-outs \( \mathbf{o}\!\sim\!\pi_\theta \) but must not contain the policy parameters \( \theta \) explicitly, exactly the premise of the REINFORCE identity \citep{sutton1998reinforcement}.  
(ii) \emph{Finite reward}: values must be finite; GRPO’s subsequent advantages  normalization already stabilizes variance, so no extra clipping is required.  
(iii) \emph{\(\theta\)-independent reward noise}: each component \( \tilde R_i(\mathbf{o}) \) can be deterministic or can include additional randomness, provided that randomness is drawn independently of \( \theta \); this keeps the likelihood-ratio estimator unbiased \citep{Williams1992}.  
When the task already ranks completions (e.g.\ correct \({>}\) incorrect), any shaping term should preserve that order; for example potential-based shaping \citep{Ng1999PolicyInvariance} provides this guarantee.  
The frequency-aware reward of in sec. \ref{sec:freq-reward} satisfies all three conditions and behaves as an entropy bonus that links GAPO to maximum-entropy RL.

\subsection{Group-Based Reward for Uniform Sampling}
\label{sec:freq-reward}

To promote output diversity while ensuring validity, we design a simple frequency-aware reward that encourages uniform sampling over a predefined set of valid responses. This leverages GAPO’s group-level view to penalize over-represented outputs and favour under-represented ones.

\paragraph{Setup}  
Let \( \mathcal{V} = \{v_1,\dots,v_L\} \) be the set of valid outputs, and let \( \mathbf{o} = (o_1,\dots,o_G) \) denote a group of rollouts. Each \( o_i \) is either a valid item in \( \mathcal{V} \) or an invalid response.

\paragraph{Frequency-Aware Reward}  
The empirical frequency of each valid item \( v \) is
\[
f_v(\mathbf{o}) = \frac{\sum_{i=1}^{G}\mathbf{1}\{o_i=v\}}
                       {\sum_{i=1}^{G}\mathbf{1}\{o_i\in\mathcal{V}\}}.
\]
Assuming a uniform target distribution \( u = 1/L \), the reward for rollout \( i \) is
\[
\tilde{R}(\mathbf{o})_i =
\begin{cases}
1 -\bigl(f_{o_i} - \frac{1}{L}\bigr), & o_i \in \mathcal{V},\\[4pt]
-1, & \text{otherwise}.
\end{cases}
\]

This design rewards under-represented valid items and penalises frequent ones, encouraging the policy to spread probability mass evenly across \( \mathcal{V} \).  The resulting vector \( \tilde{R}(\mathbf{o}) \) is fed directly into the GRPO update.

\section{Experiments}
We trained models from the Qwen2.5 Instruct family~\cite{yang2024qwen2} using our proposed GAPO method with the frequency-aware reward function introduced in Section~\ref{sec:gapo}.
The models were fine-tuned using LoRA~\cite{hu2022lora}. 
For training, we constructed a synthetic dataset comprising random lists from diverse topics, with list lengths ranging from 4-12 items (see examples in Appendix~\ref{app:app_training_examples}, and additional implementation details in Appendix \ref{app:app_implementation_details}). 
In these experiments, the models were instructed to sample a single item from each list. 
\begin{figure*}[!htbp]
  \centering
  \makebox[\textwidth][c]{
    \begin{subfigure}[b]{0.32\linewidth}
      {\footnotesize
      \begin{tcolorbox}[
        width=\linewidth,
        height=15mm,
        colback=black!5,
        colframe=black,
        arc=2pt,
        boxrule=1pt,
        left=1pt, right=1pt, top=1pt, bottom=1pt,
        title=\textbf{Instruction},
        fonttitle={\bfseries\color{white}},
        colbacktitle=black,
        coltitle=white,
      ]
      Name one city anywhere in the world...
      \end{tcolorbox}}
      \includegraphics[width=\linewidth]{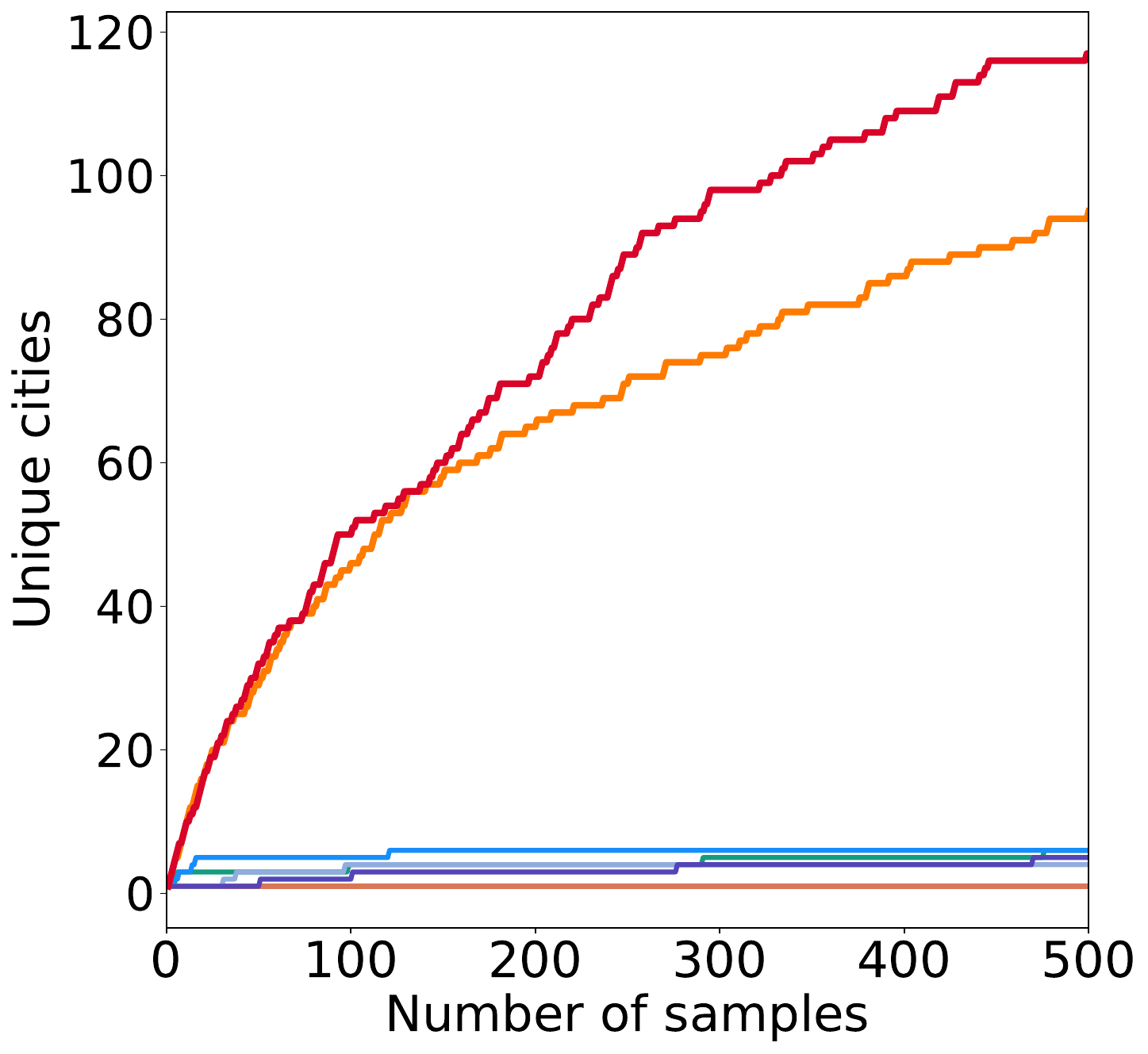}
      \tiny \caption{Cities.}
      \label{fig:plot2}
    \end{subfigure}
    \hfill
    \begin{subfigure}[b]{0.32\linewidth}
      {\footnotesize
      \begin{tcolorbox}[
        width=\linewidth,
        height=15mm,
        colback=black!5,
        colframe=black,
        arc=2pt,
        boxrule=1pt,
        left=1pt, right=1pt, top=1pt, bottom=1pt,
        title=\textbf{Instruction},
        fonttitle={\bfseries\color{white}},
        colbacktitle=black,
        coltitle=white,
      ]
      Provide the name of a random celebrity...
      \end{tcolorbox}}
      \includegraphics[width=\linewidth]{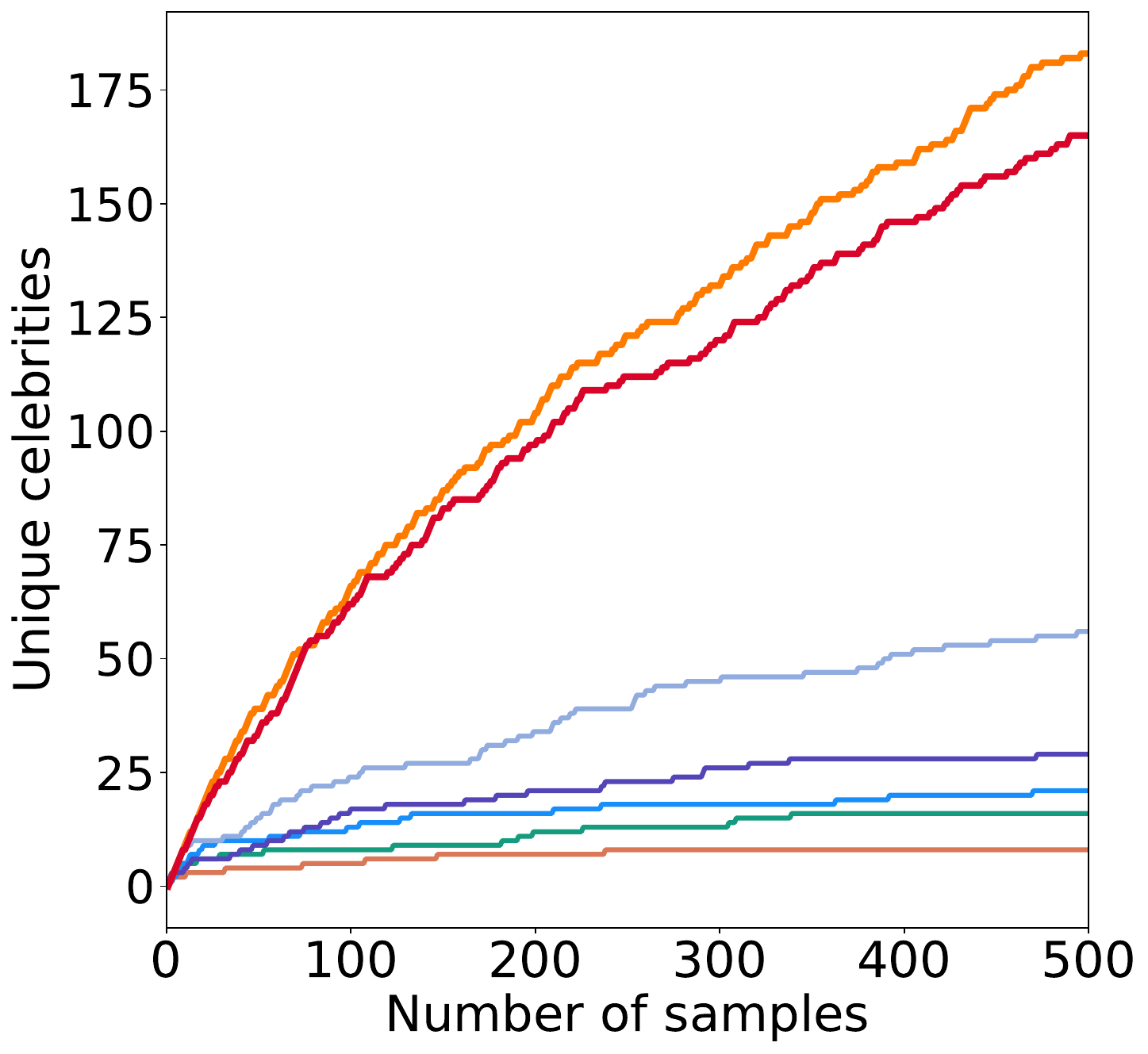}
      \tiny \caption{Celebrities.}
      \label{fig:plot4}
    \end{subfigure}
    \hfill
    \begin{subfigure}[b]{0.32\linewidth}
      \includegraphics[width=\linewidth]{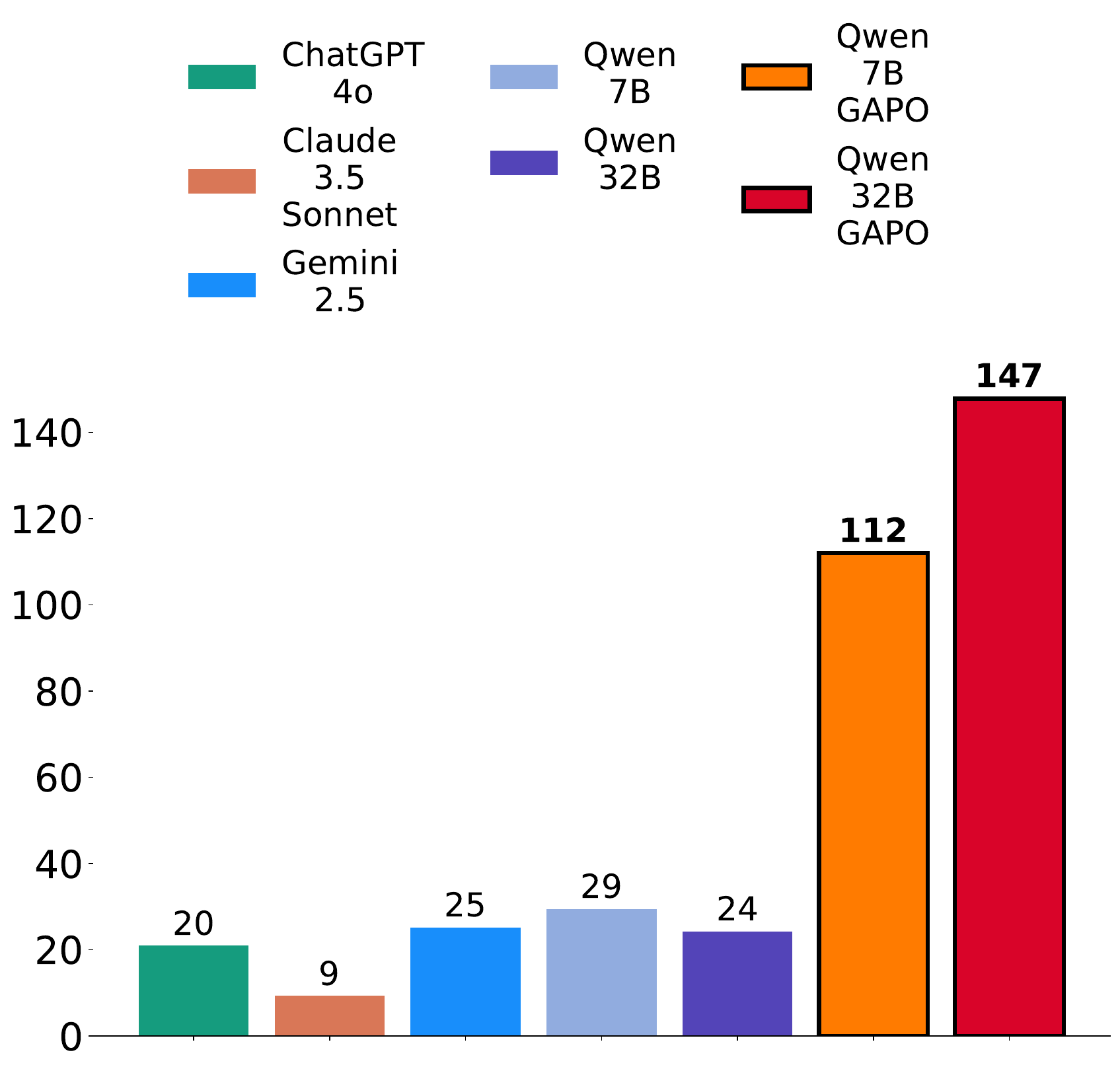}
      \tiny \caption{Avg. Unique@500 ($\uparrow$).}
      \label{fig:plot3}
    \end{subfigure}
    \hfill}
  \caption{\textbf{Open-set diversity.} (a,b) Cumulative number of unique responses across 500 samples for open-ended prompts. (c) Average number of unique responses across 500 samples, aggregated over all ten categories.}
  \vspace{-1em}
  \label{fig:open_set}
  
\end{figure*}

\subsection{Uniformity Experiments}
We first evaluate our approach on a task directly aligned with our objective: sampling items uniformly from a fixed list. 
For this experiment, we constructed 10 distinct lists, each containing eight items from different categories (\eg, planets, musical instruments, books), and issued identical selection prompts 100 times per model and list. Importantly, these categories were not seen during GAPO training.

Figure~\ref{fig:planets_close_set} and \ref{fig:music_close_set} present the distributions of model responses for planets and musical instruments across all models: ChatGPT-4o, Claude Sonnet 3.5, Gemini 2.5 Flash, and Qwen2.5 Instruct (7B/32B), with the latter shown both before and after GAPO fine-tuning. 
To quantify uniformity, Figure~\ref{fig:avg_js_close_set} shows the Jensen-Shannon divergence (JSD) computed between each model's empirical distribution and the ideal uniform distribution over valid items (\ie, $12.5\%$ each, with $0\%$ for invalid outputs). JSD is preferred over Kullback-Leibler divergence here, as it remains defined even when distributions have non-overlapping support.

As shown in Figure~\ref{fig:avg_js_close_set}, GAPO-trained models consistently achieve significantly lower divergence (JSD $< 0.1$) compared to all baselines (JSD $> 0.3$), indicating distributions much closer to uniform. Complete results across all 10 categories are provided in Appendix~\ref{app:app_uniformity}. The visualizations confirm that GAPO-trained models produce distributions substantially closer to uniform, while baseline models consistently over-represent certain choices.

\subsection{Open Questions Experiments}
In this experiment, we ask the model to return a single item belonging to a specific category, \eg, \textit{``Name one city anywhere in the world''}, without providing a list of options.
We randomly selected 10 categories and ask each model to name an item from the category 500 times while counting the unique items each model presents.
This task is both more complicated than selecting an item from a list and differs from the training objective for showing generalization.
Figure~\ref{fig:open_set} shows that all regular models sample only a few different items per task, while our models sample many more unique items.
For example, our finetuned Qwen2.5 32B samples on average 147 unique items compared to 24 sampled by Qwen2.5 32B before finetuning.
The eight categories not presented in Figure~\ref{fig:open_set} are presented in Appendix~\ref{app:open_set}.

\begin{table*}
\centering
\caption{Comparison of diversity metrics across creative writing tasks. GAPO consistently outperforms the baseline model in both semantic diversity (higher embedding distances) and lexical uniqueness (higher 1-Self-BLEU scores).}
\vspace{-0.6em}
\label{tab:text_creativity}
\resizebox{\textwidth}{!}{%
\begin{tabular}{l|cc|cc}
\toprule
\multirow{2}{*}[-0.5ex]{\textbf{Creative Writing Task}} & \multicolumn{2}{c|}{\textbf{Avg. Embedding Distance} ($\uparrow$)} & \multicolumn{2}{c}{\textbf{Avg.} $\mathbf{1-}$\textbf{Self-BLEU} ($\uparrow$)} \\
\cmidrule{2-5}
 & \multicolumn{1}{c}{\hspace{0.4cm}\textbf{Baseline}} & \multicolumn{1}{c|}{\hspace{0.3cm}\textbf{GAPO (ours)}} & \multicolumn{1}{c}{\textbf{Baseline}} & \multicolumn{1}{c}{\textbf{GAPO (ours)}} \\
\midrule
Write a story with no more than 100 words & 0.31 & \textbf{0.44} & 0.83 & \textbf{0.95} \\
Write a poem with no more than 100 words & 0.17 & \textbf{0.20} & 0.73 & \textbf{0.93} \\
Write a haiku in English & 0.1 & \textbf{0.21} & 0.54 & \textbf{0.80} \\
Craft a one-sentence mystery opening & 0.40 & \textbf{0.59} & 0.67 & \textbf{0.93} \\
Compose a two-line dialogue between two characters & 0.21 & \textbf{0.57} & 0.58 & \textbf{0.85} \\
Pitch an idea for a new fruit in one sentence & 0.01 & \textbf{0.44} & 0.16 & \textbf{0.78} \\
Tell a joke & 0.15 & \textbf{0.37} & 0.19 & \textbf{0.37} \\
Write only the chorus for a pop song & 0.25 & \textbf{0.43} & 0.65 & \textbf{0.94} \\
\midrule
\textbf{Average} & 0.20 & \textbf{0.41} & 0.54 & \textbf{0.82} \\
\bottomrule
\end{tabular}
}
\vspace{-0.5em}
\end{table*}
\subsection{Creativity}
\label{sec:creativity}
\begin{figure}[H]
\centering
  \includegraphics[width=0.85\linewidth]{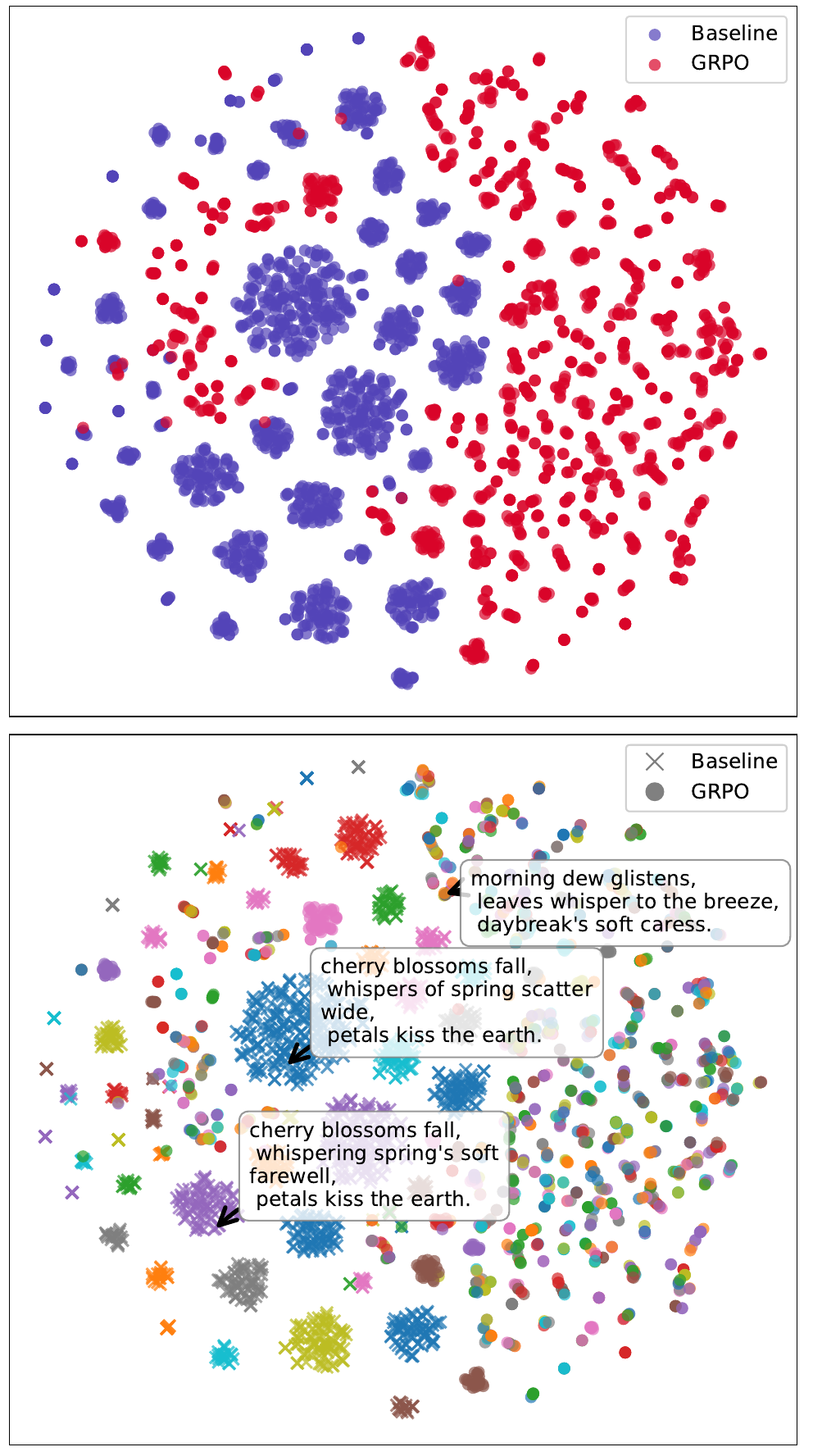}
  \caption{\textbf{Creativity visualization.} t-SNE Visualization of embeddings of responses for the prompt \textit{``Write a haiku in English.''}.
  In the bottom plot, identical haikus or text responses are represented with the same color. Small random noise was added to spread identical responses into visible clusters.}
  \label{fig:creativity}
  \vspace{-1em}
\end{figure}

Diversified model outputs are particularly important for creative writing tasks. To assess our method's output diversity in creative writing scenarios, we conducted two experiments.


First, we generated 1500 short haikus using the prompt \textit{``Write a haiku in English.''} with both the baseline and GAPO-trained Qwen2.5 32B Instruct models. 
We then embedded each story using a Transformer-based embedding model~\cite{embedding_model_mpnet} and visualized the embedding space using t-SNE. 
Figure~\ref{fig:creativity} demonstrates that the GAPO model produces haikus that occupy a substantially broader region of the embedding space.
This indicates that GAPO generates haikus with greater diversity compared to the baseline model.

Next, to quantify output diversity across a wider range of creative tasks, we prompted each model with eight different  writing instructions and generated 100 outputs for each prompt. We then computed two complementary diversity metrics: (1) average embedding distance between all pairs of outputs, which captures semantic diversity, and (2) average 1 - Self-BLEU score \cite{self-bleu, papineni2002bleu}, which measures lexical diversity. Table~\ref{tab:text_creativity} presents these results.

As shown in Table~\ref{tab:text_creativity}, the GAPO model consistently produces more diverse outputs across all creative tasks. 
On average, GAPO improves the embedding distance by 160\% (from 0.15 to 0.39) and the 1-Self-BLEU score by 75\% (from 0.52 to 0.91). The improvements are particularly notable for tasks like \textit{``Compose a two-line dialogue''} and \textit{``Tell a joke''}, where the baseline model shows near-zero diversity (indicating almost identical outputs), while GAPO achieves substantial variation.

Finally, to qualitatively illustrate our method's superior output diversity compared to the baseline, side-by-side comparisons of responses generated by each model on several of the aforementioned creative writing tasks are presented in Appendix~\ref{app:responses_comp}. These results verify that GAPO effectively promotes output diversity in open-ended creative writing tasks compared to the baseline.
\begin{table*}[t]
  \centering
  \caption{Performance comparison of Qwen2.5 32B Instruct Baseline and GAPO-trained models on standard benchmarks at temperature 0.7. GAPO improves flexible scoring and output diversity while maintaining or improving generalization.}
  \label{tab:accuracy-results}
  \vspace{-0.5em}
  \small
  \begin{tabular}{c|cc|cc|c|c}
    \toprule
    \multirow{2}{*}{\textbf{Model}} & \multicolumn{2}{c|}{\textbf{GSM8K}} & \multicolumn{2}{c|}{\textbf{MATH}} & \multirow{2}{*}{\textbf{HumanEval}} & \multirow{2}{*}{\textbf{MMLU-Pro}} \\
    & Exact & Flexible & Verify & Exact Match & & \\
    \midrule
    Baseline & \textbf{0.835} & 0.865 & 0.484 & \textbf{0.524} & 0.555 & \textbf{0.675} \\
    GAPO (ours) & 0.772 & \textbf{0.905} & \textbf{0.499} & 0.502 & \textbf{0.579} & 0.656 \\
    \bottomrule
  \end{tabular}
  \vspace{-0.5em}
\end{table*}

\subsection{Benchmarks}
\label{sec:benchmarks}


While increasing diversity is valuable, a key concern is that it may come at the expense of accuracy or coherence. We assess GAPO on standard reasoning and knowledge benchmarks to verify it remains competitive with the baseline. This evaluation is critical because diversity-promoting methods could potentially interfere with the model's ability to produce correct, coherent responses in tasks that require precise reasoning or factual accuracy.
We report results on \textasciitilde200 sample subsets of the following benchmarks:

\begin{itemize}
    \setlength\parskip{0pt}
    \setlength\topsep{0pt}
    \item \textbf{GSM8K}~\cite{gsm8k_paper}, a dataset of grade-school math problems requiring multi-step reasoning. 
    \item \textbf{MATH}~\cite{hendrycksmath2021}, a benchmark of advanced mathematical problems. 
    \item \textbf{HumanEval}~\cite{chen2021evaluatinglargelanguagemodels}, a code generation benchmark assessing functional correctness.
    \item \textbf{MMLU-Pro}~\cite{wang2024mmlu}, a multi-choice exam of diverse domains.
\end{itemize}


To ensure robust results, we report the average performance across five evaluation runs for each subset (see Appendix ~\ref{app:app_implementation_details}).
\\
Table~\ref{tab:accuracy-results} presents results for Qwen2.5 32B Instruct at temperature 0.7. GAPO performs comparably to the baseline across all benchmarks.
These results suggest that GAPO can improve output diversity while maintaining similar performance levels to the original model.

\begin{figure}[htb]
  \includegraphics[width=0.95\linewidth]{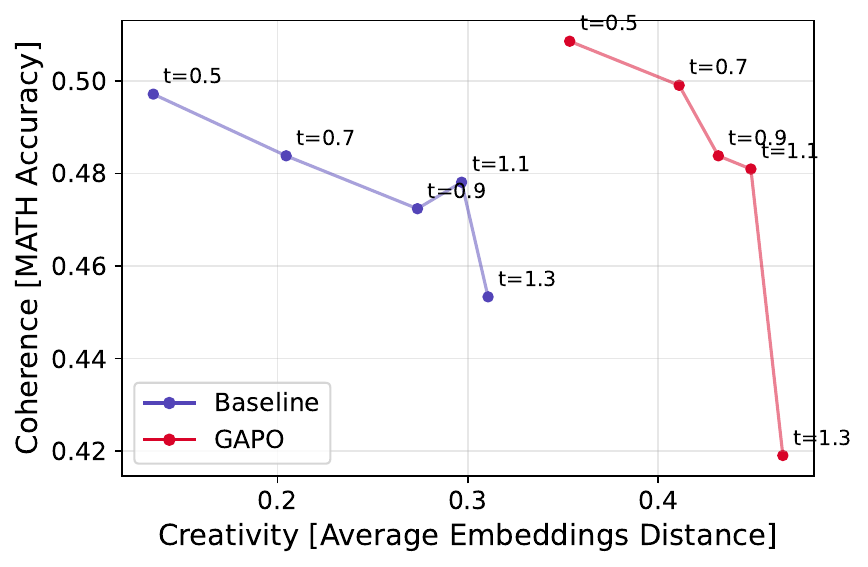}
 \caption{\textbf{Creativity vs. Coherence tradeoff.} Comparison of mathematical reasoning accuracy (MATH) against creativity (embeddings cosine distance) for GAPO and Baseline models across sampling temperatures (t=0.5-1.3).}
  \label{fig:accuracy_vs_creativity}
 \vspace{-1em}
\end{figure}




\subsection{Creativity-Coherence Tradeoff}
\label{sec:creativity_coherence_tradeoff}
We have empirically validated that our GAPO model is more creative than the baseline. In this section, we further verify that the increased diversity does not come at the expense of coherence.

We evaluate both the baseline and GAPO Qwen2.5 32B models across multiple temperature settings, measuring coherence by accuracy on the MATH dataset and creativity by the average cosine distance between response embeddings, computed as described in Section~\ref{sec:creativity}. The creative writing prompts used for this analysis are listed in Appendix~\ref{app:creativity_coherence}. 
As shown in Figure~\ref{fig:accuracy_vs_creativity}, GAPO consistently achieves higher creativity at each coherence level, indicating improved diversity without loss of coherence.

\section{Limitations}
Our work has several key limitations. We focused on LoRA fine-tuning rather than full model tuning or earlier integration in the instruction pipeline. The reward function assumes equally valid completions, making it best suited for list selection and harder to extend to accuracy-diversity tradeoffs.

Finally, while we show generalization to unseen categories and open-ended questions, the limits of this generalization are not fully understood.

\paragraph{Potential Risks} GAPO may generate broader ranges of problematic content due to increased diversity, though base model safety properties should be preserved. The diversity-accuracy tradeoff could impact safety-critical applications, requiring task-specific evaluation before deployment. Our synthetic dataset contains potential biases that may propagate to trained models. Additionally, the computational requirements may limit access for smaller research groups.
\section{Conclusions}
We introduced \textit{Group-Aware Policy Optimization} (GAPO), a simple extension of GRPO that computes rewards over groups of completions instead of individual samples. This group-based reward formulation enables training for distributional properties such as diversity and uniform coverage.
Combined with our frequency-aware reward function, GAPO effectively counters mode collapse in LLMs, producing near-uniform distributions on list selection tasks and improving diversity and creativity on open-ended prompts, without sacrificing coherence.

Future work should explore integrating GAPO earlier in the training pipeline and extending its reward functions to balance diversity with task-specific accuracy, enabling its application to open-ended tasks where the space of valid responses is implicit or unbounded.

\newpage




\bibliographystyle{acl_natbib}
\bibliography{main}
\newpage

\onecolumn  
\appendix

\label{sec:appendix}
\section{Uniformity - Additional Results}
\label{app:app_uniformity}
\begin{figure*}[!htb]
  \centering
  \scalebox{0.92}{ 
  \begin{minipage}{\textwidth}
  \scriptsize
  \begin{tcolorbox}[
        width=0.95\textwidth,
        colback=black!5,
        colframe=black,
        arc=2pt,
        boxrule=1pt,
        left=4pt, right=4pt, top=4pt, bottom=4pt,
        title=\textbf{Instruction},
        fonttitle={\bfseries\color{white}},
        colbacktitle=black,
        coltitle=white
        ]
      Instruction: Please select one of the following items \{list\}.\\
      Format your response as follows: \texttt{<answer>selected\_item</answer>}.
  \end{tcolorbox}
  
  \begin{subfigure}[b]{0.32\textwidth}
    \includegraphics[width=\linewidth]{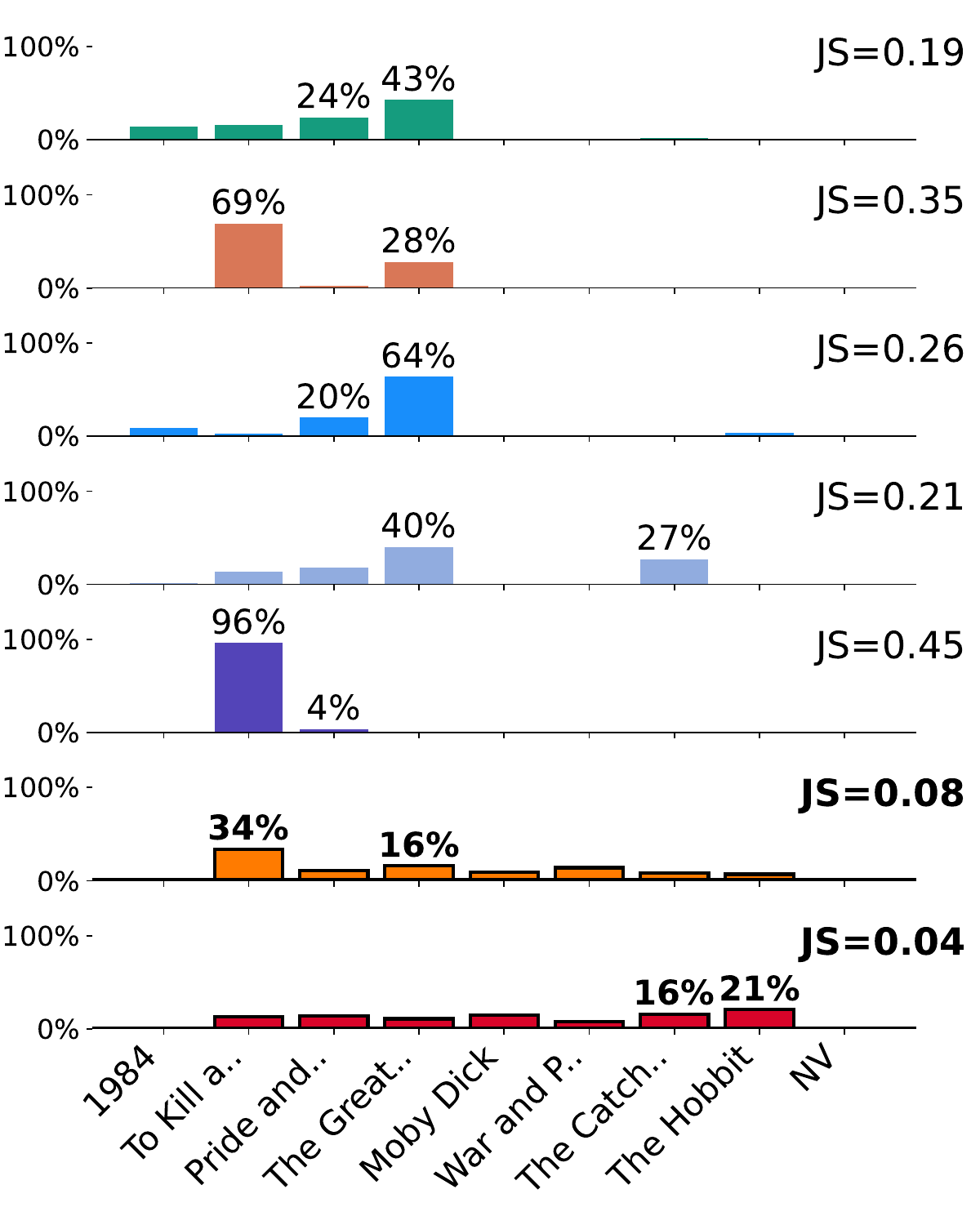}
    \caption{Books.}
    \label{fig:books_close_set}
  \end{subfigure}\hfill
  \begin{subfigure}[b]{0.32\textwidth}
    \includegraphics[width=\linewidth]{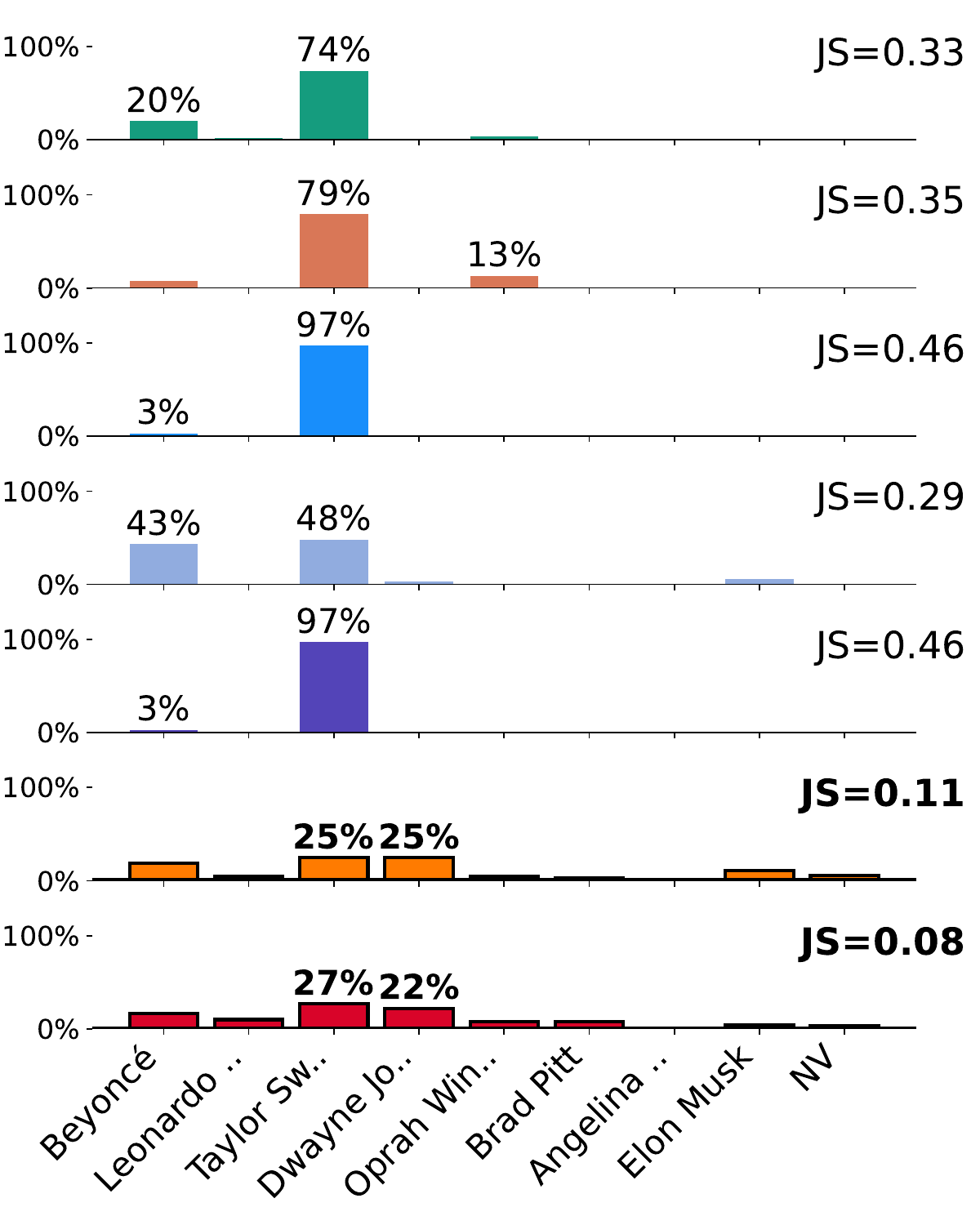}
    \caption{Celebrities.}
    \label{fig:celebrities_close_set}
  \end{subfigure}\hfill
  \begin{subfigure}[b]{0.32\textwidth}
    \includegraphics[width=\linewidth]{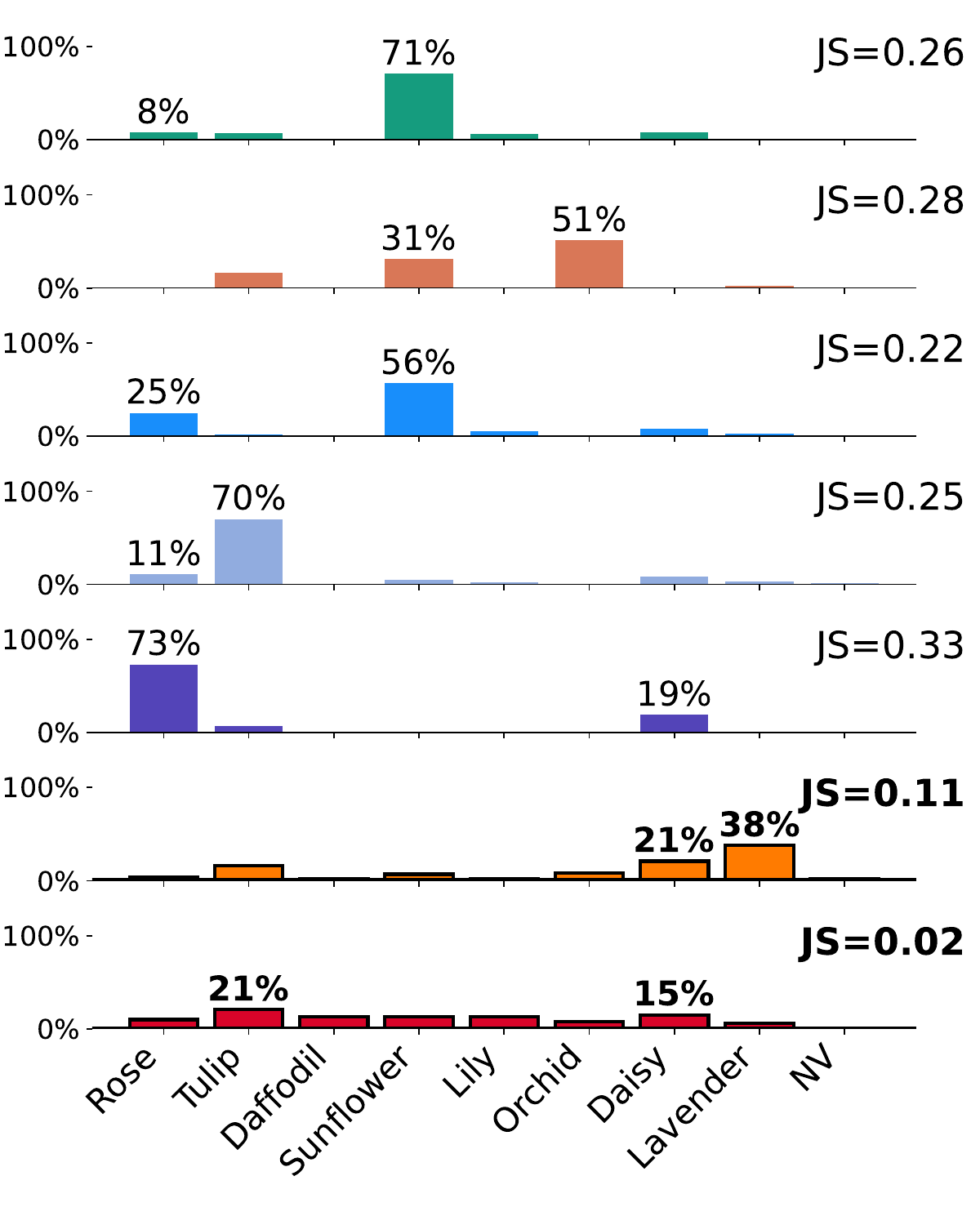}
    \caption{Flowers.}
    \label{fig:flowers_close_set}
  \end{subfigure}
  
  \begin{subfigure}[b]{0.32\textwidth}
    \includegraphics[width=\linewidth]{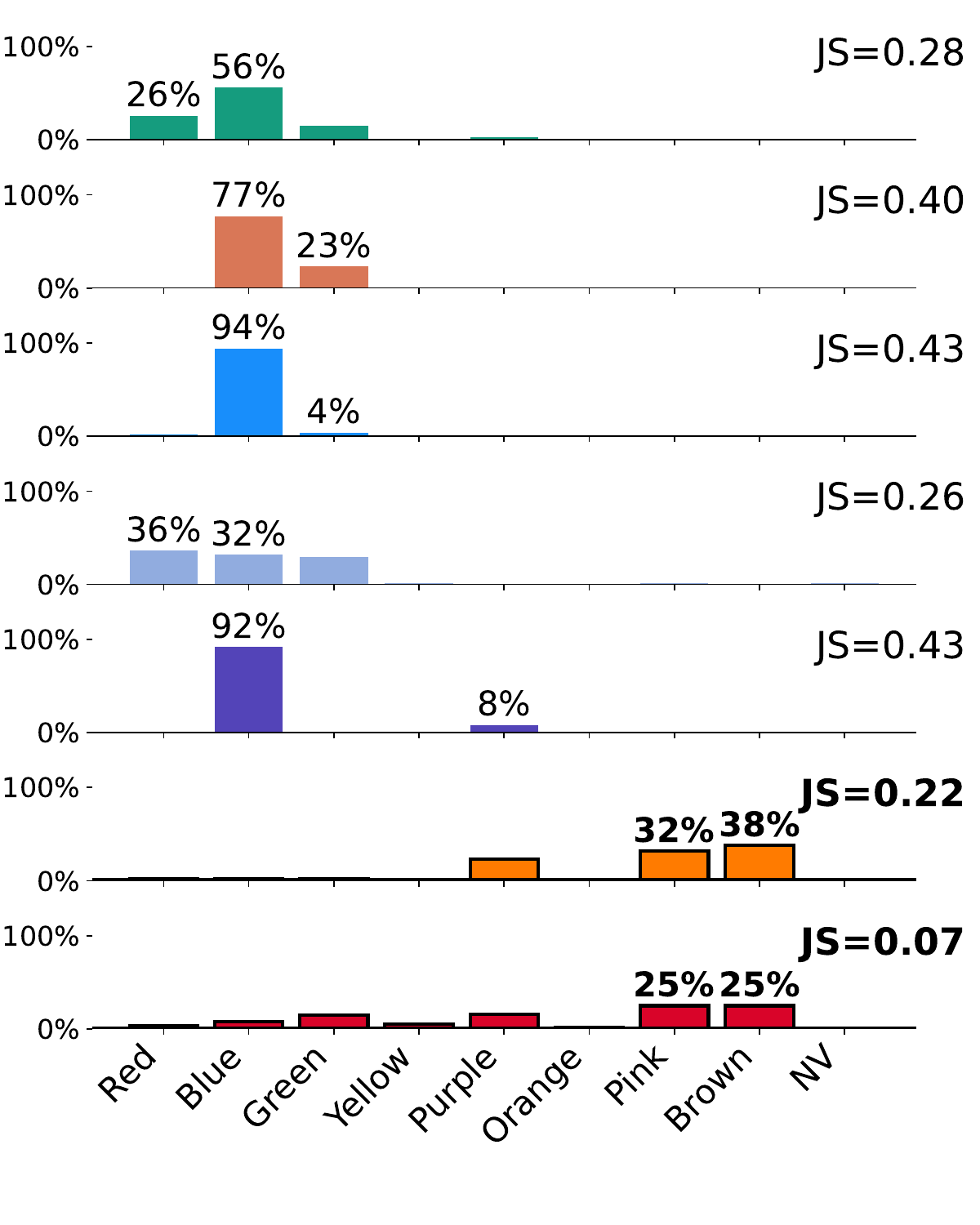}
    \caption{Colors.}
    \label{fig:colors_close_set}
  \end{subfigure}\hfill
  \begin{subfigure}[b]{0.32\textwidth}
    \includegraphics[width=\linewidth]{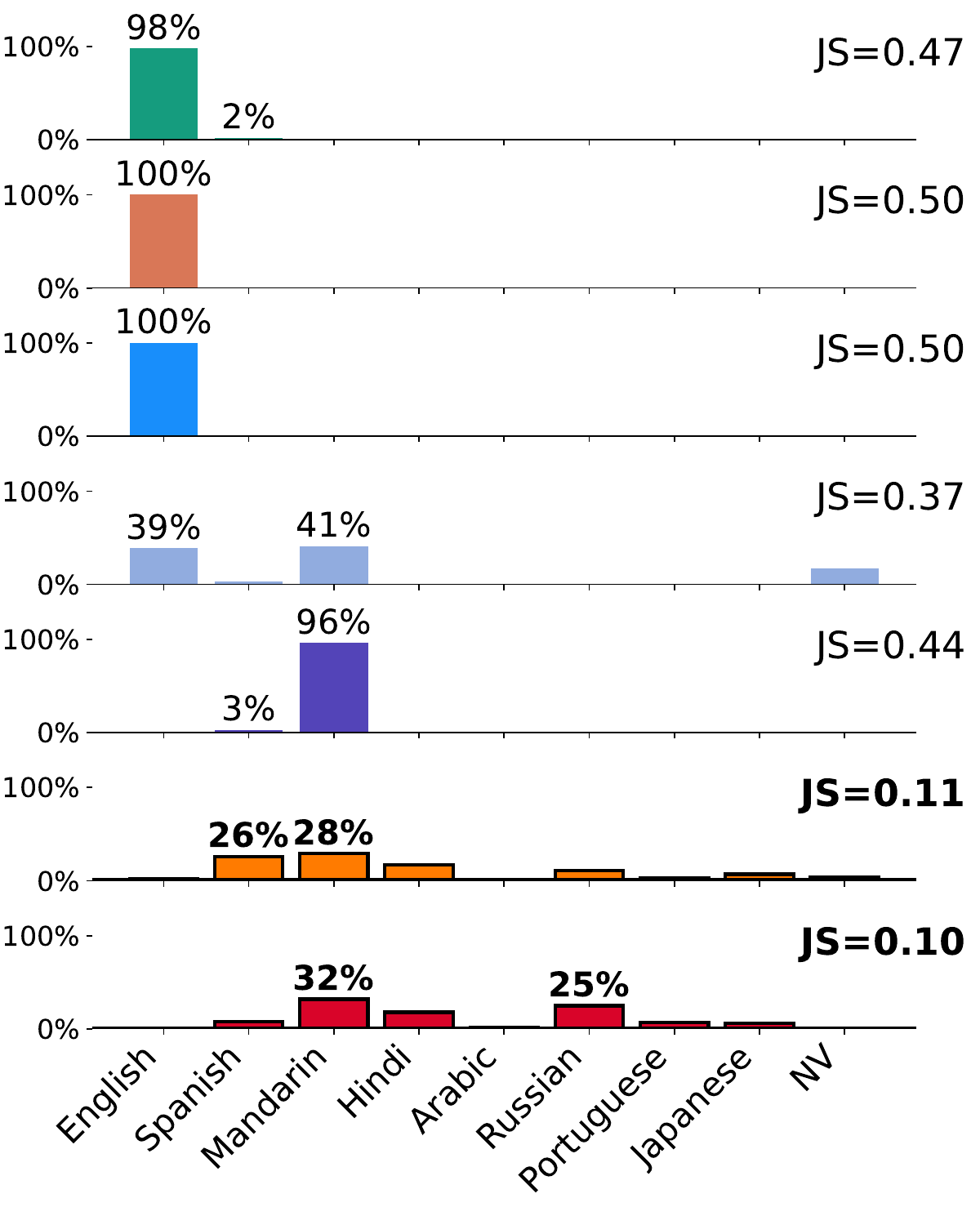}
    \caption{Languages.}
    \label{fig:languages_close_set}
  \end{subfigure}\hfill
  \begin{subfigure}[b]{0.32\textwidth}
    \includegraphics[width=\linewidth]{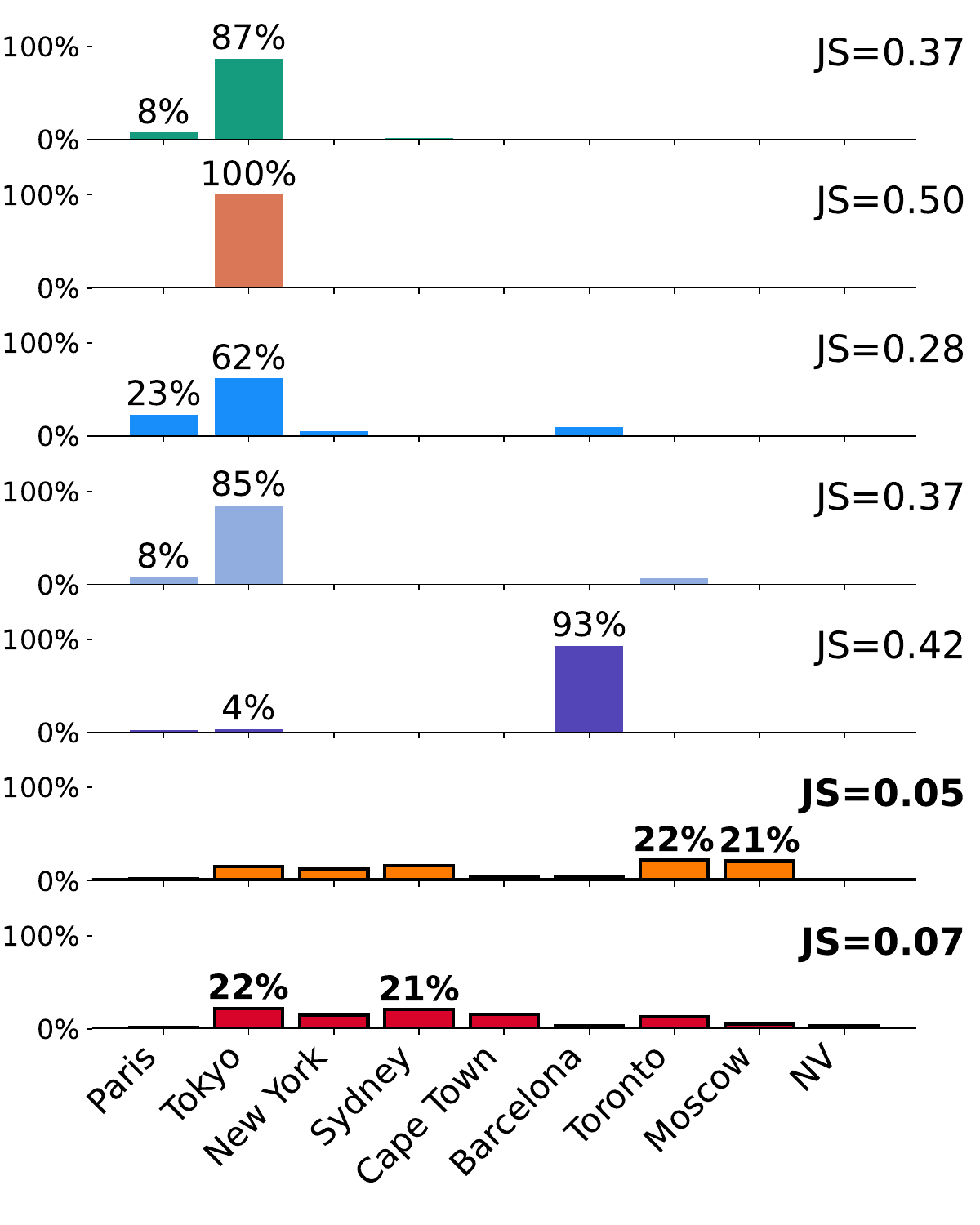}
    \caption{Cities.}
    \label{fig:cities_close_set}
  \end{subfigure}
  
  \begin{subfigure}[b]{0.32\textwidth}
    \includegraphics[width=\linewidth]{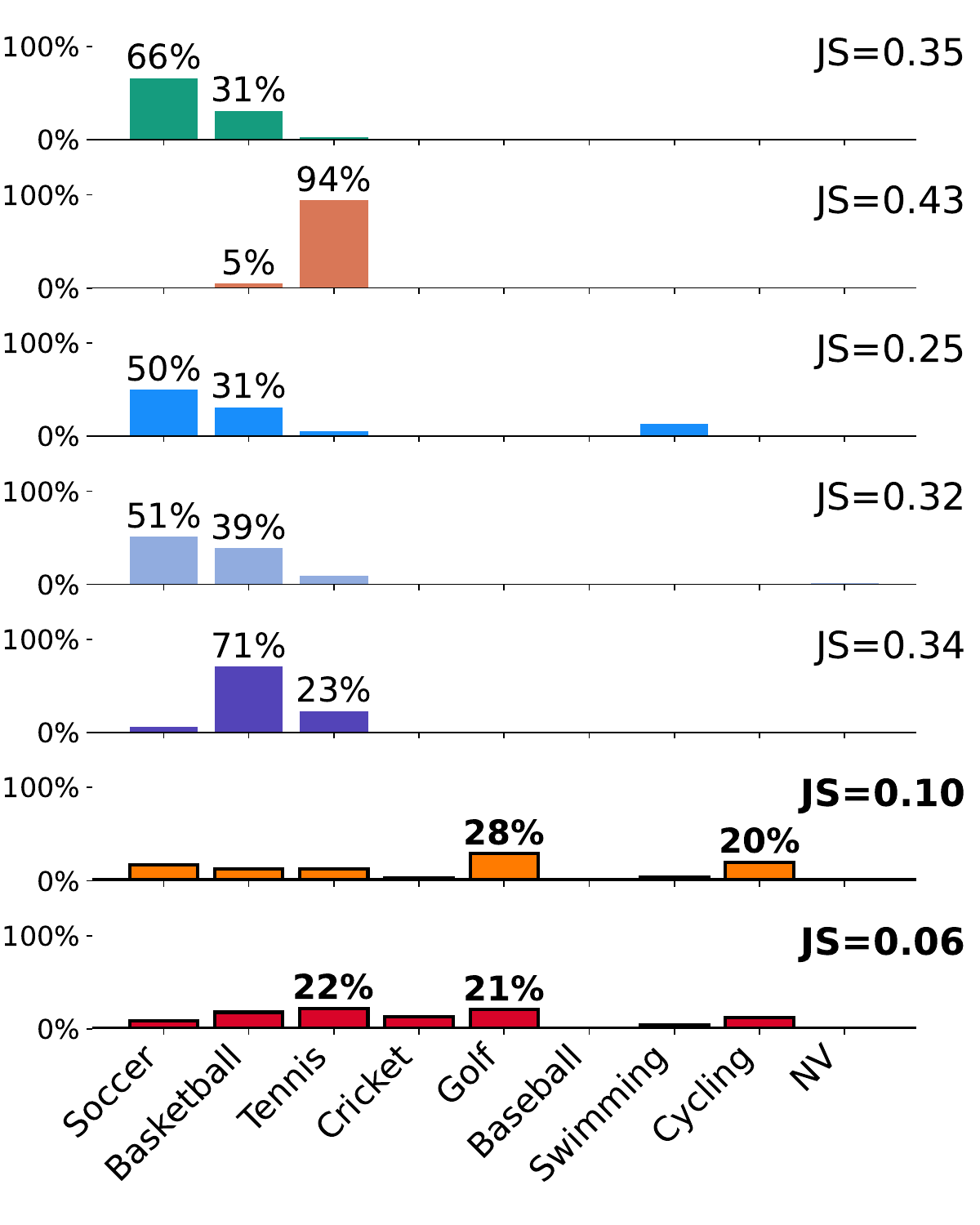}
    \caption{Sports.}
    \label{fig:sports_close_set}
  \end{subfigure}\hfill
  \begin{subfigure}[b]{0.32\textwidth}
    \includegraphics[width=\linewidth]{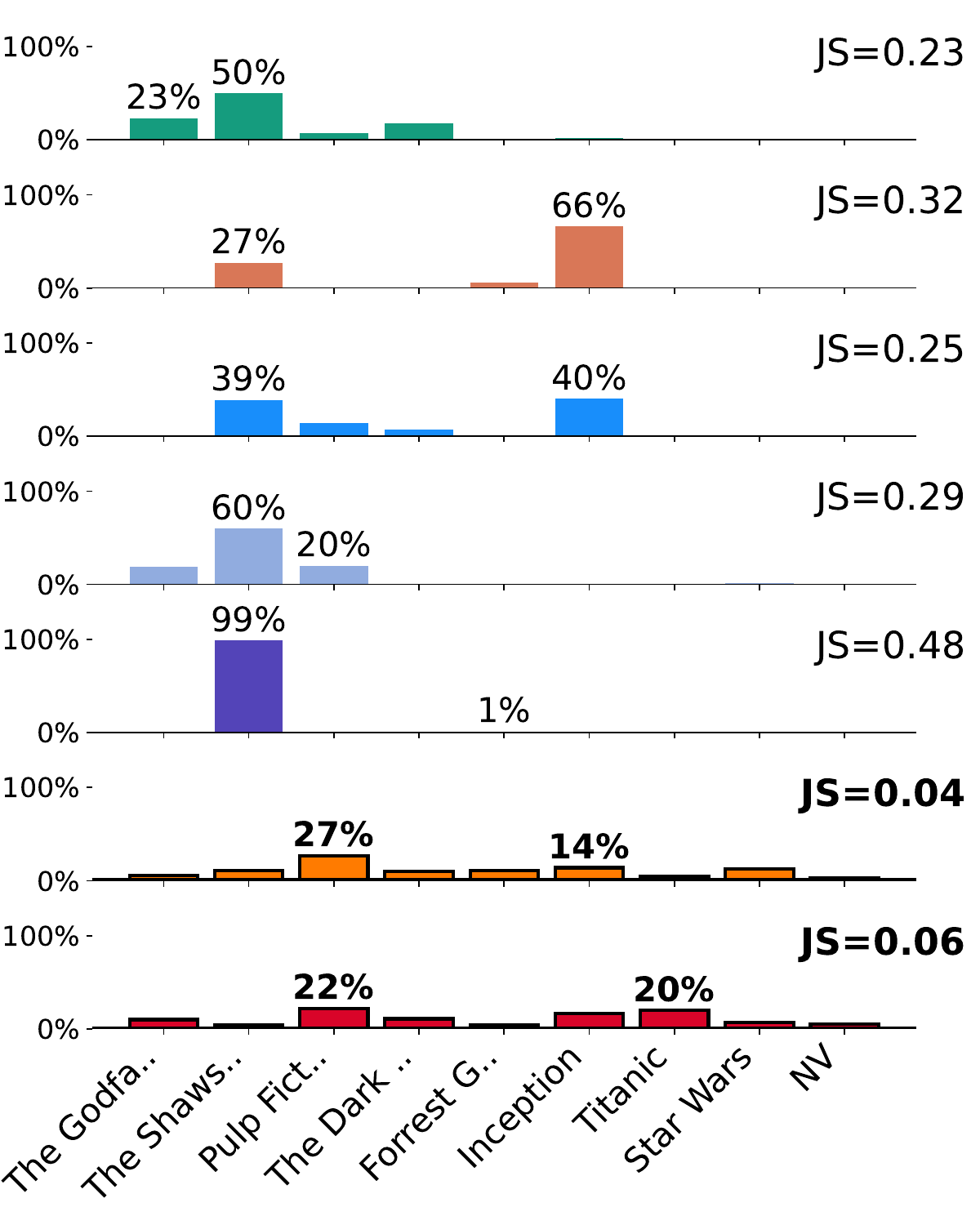}
    \caption{Movies.}
    \label{fig:movies_close_set}
  \end{subfigure}\hfill
  \begin{subfigure}[b]{0.32\textwidth}
    \includegraphics[width=\linewidth]{images/close_set/js_per_model.pdf}
    \caption{Average JS.}
    \label{fig:js_per_model}
  \end{subfigure}
  
  \caption{\textbf{Sampling from a list (Additional Results).} Comparing distribution of selections across different models when prompted to choose from a list of Books (a), Celebrities (b), Flowers (c), Colors (d), Languages (e), Cities (f), Sports (g), Movies (h). Panel (i) shows average
Jensen-Shannon divergence across all topics.}
  \label{fig:list_uniformity_appendix}
  \end{minipage}
  } 
\end{figure*}
\section{Sampling - Additional Results}
\label{app:open_set}
\begin{figure*}[!htbp]
  \centering
  \scalebox{0.92}{ 
  \begin{minipage}{\textwidth}
  \begin{subfigure}[b]{0.32\textwidth}
    \includegraphics[width=\linewidth]{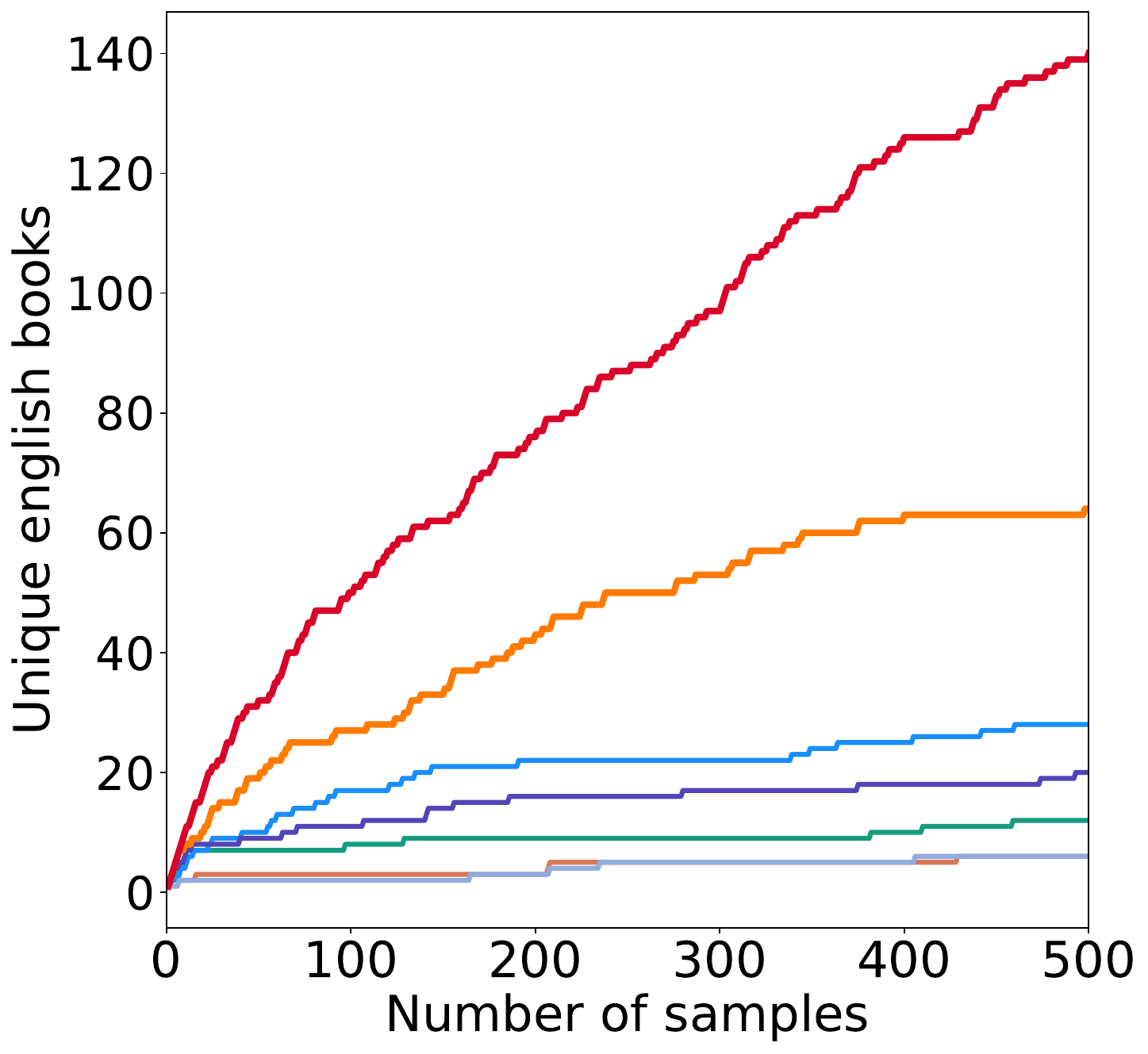}
    \caption{Books.}
    \label{fig:aaa}
  \end{subfigure}\hfill
  \begin{subfigure}[b]{0.32\textwidth}
    \includegraphics[width=\linewidth]{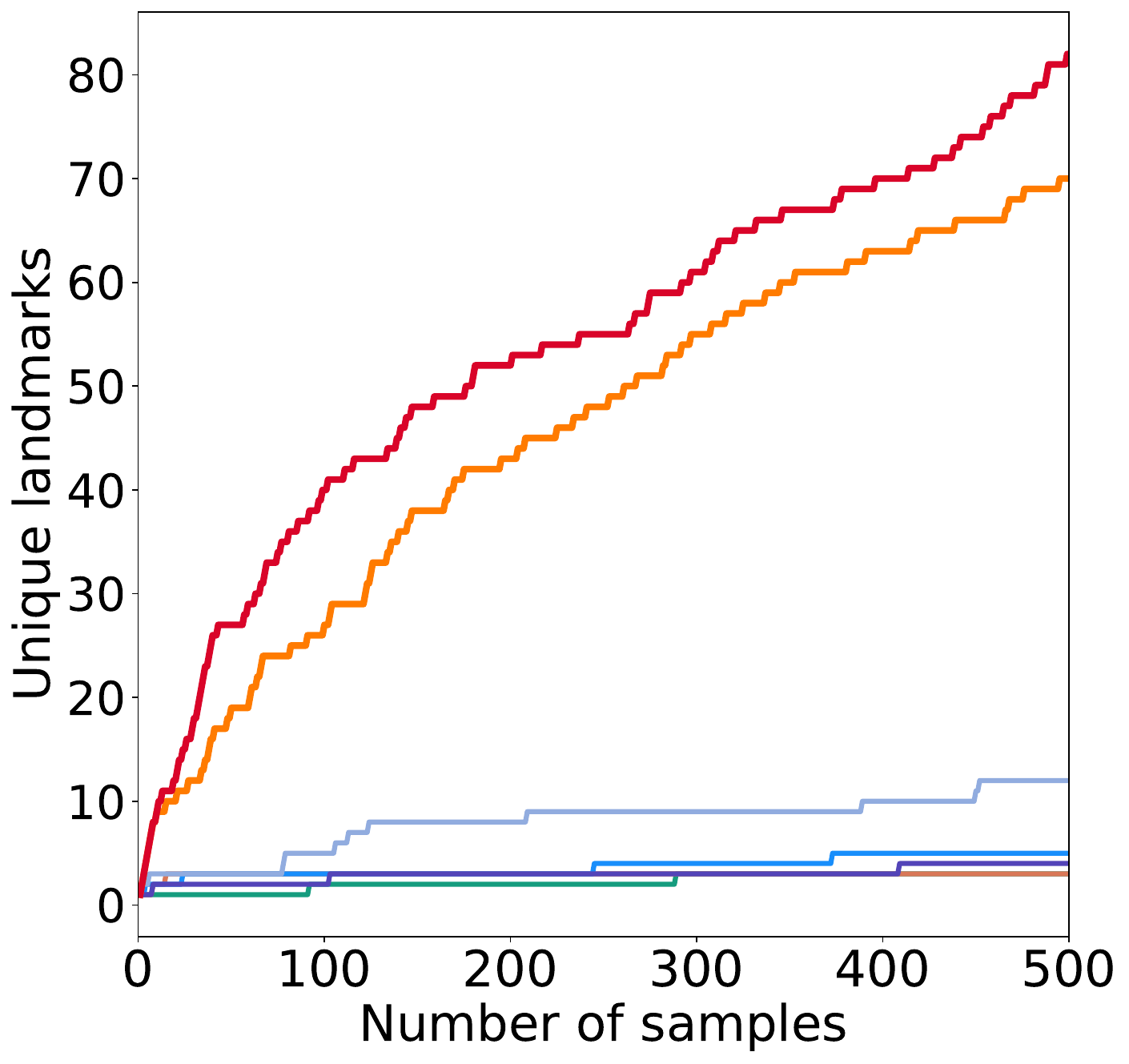}
    \caption{Landmarks.}
    \label{fig:bbb}
  \end{subfigure}\hfill
  \begin{subfigure}[b]{0.32\textwidth}
    \includegraphics[width=\linewidth]{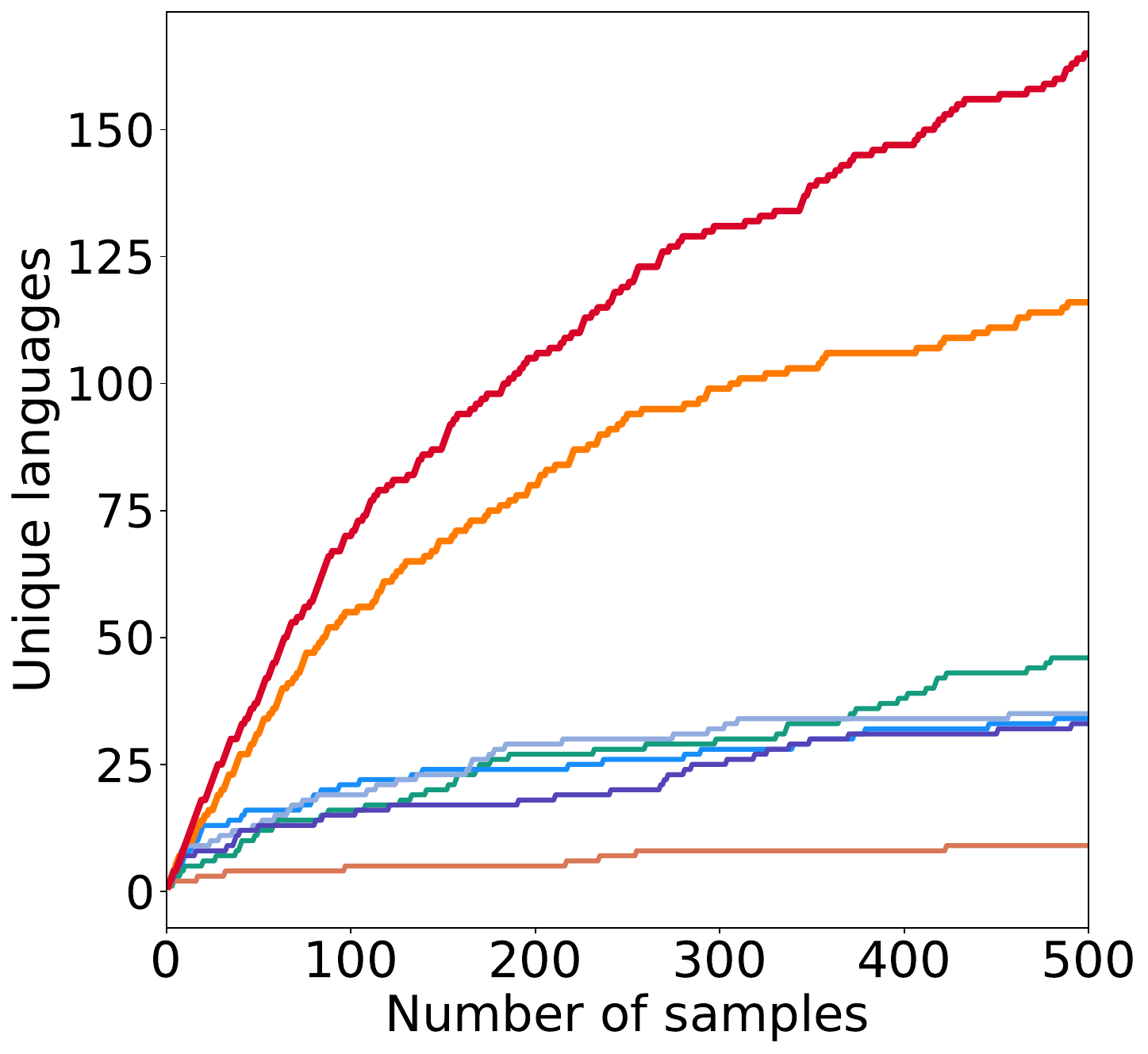}
    \caption{Languages.}
    \label{fig:ccc}
  \end{subfigure}
  
  \begin{subfigure}[b]{0.32\textwidth}
    \includegraphics[width=\linewidth]{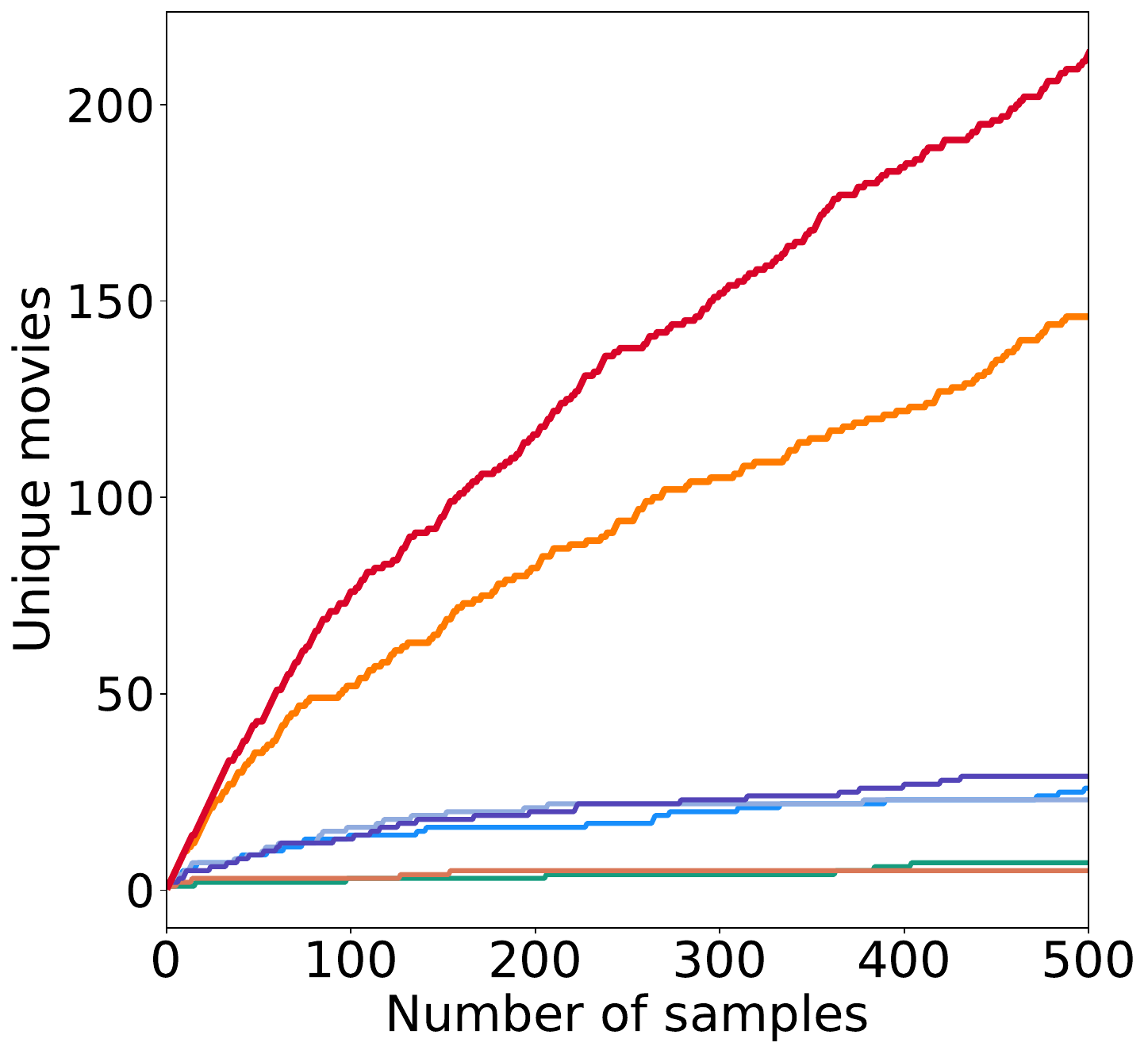}
    \caption{Movies.}
    \label{fig:ddd}
  \end{subfigure}\hfill
  \begin{subfigure}[b]{0.32\textwidth}
    \includegraphics[width=\linewidth]{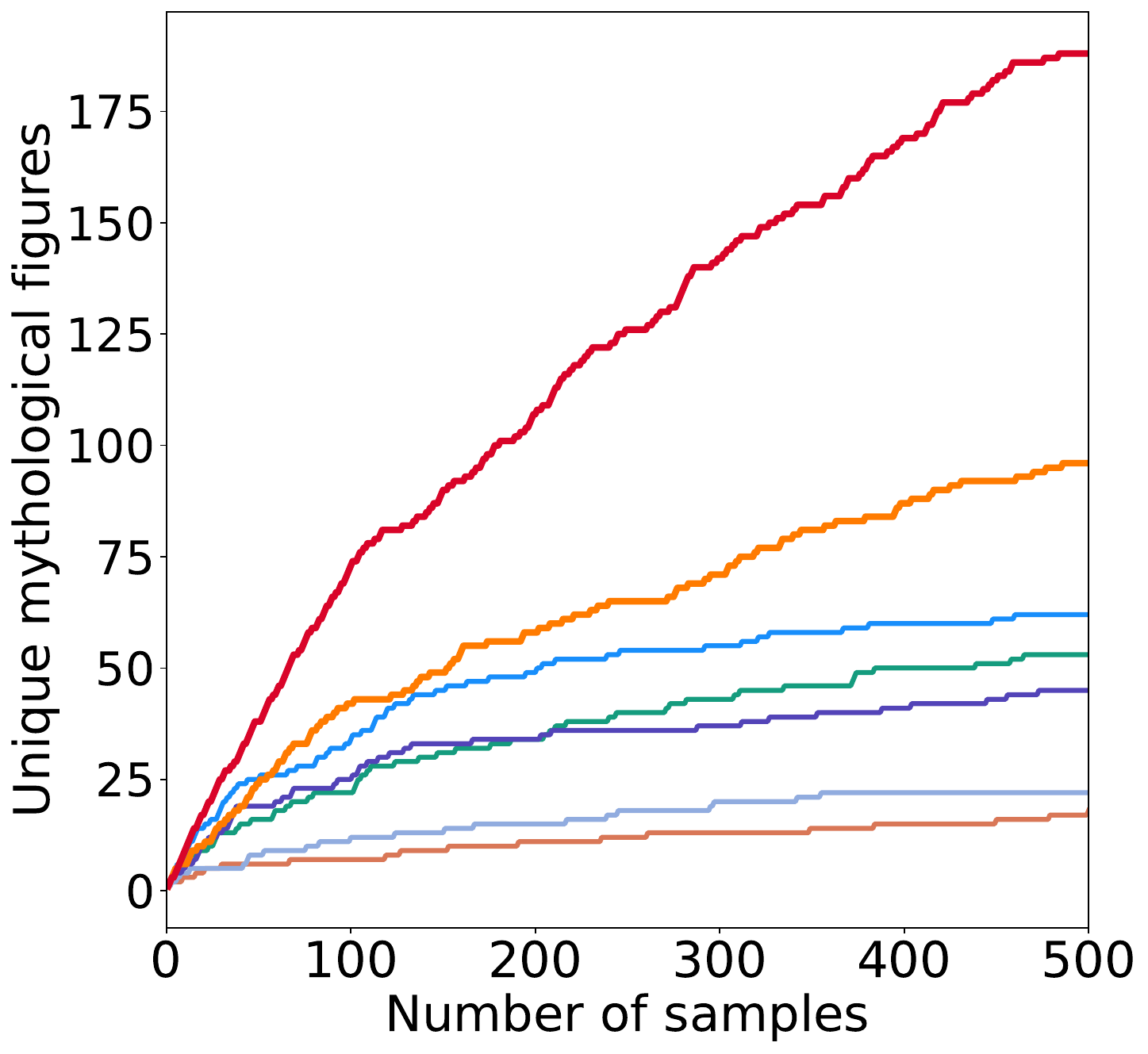}
    \caption{Mythological Figures.}
    \label{fig:eee}
  \end{subfigure}\hfill
  \begin{subfigure}[b]{0.32\textwidth}
    \includegraphics[width=\linewidth]{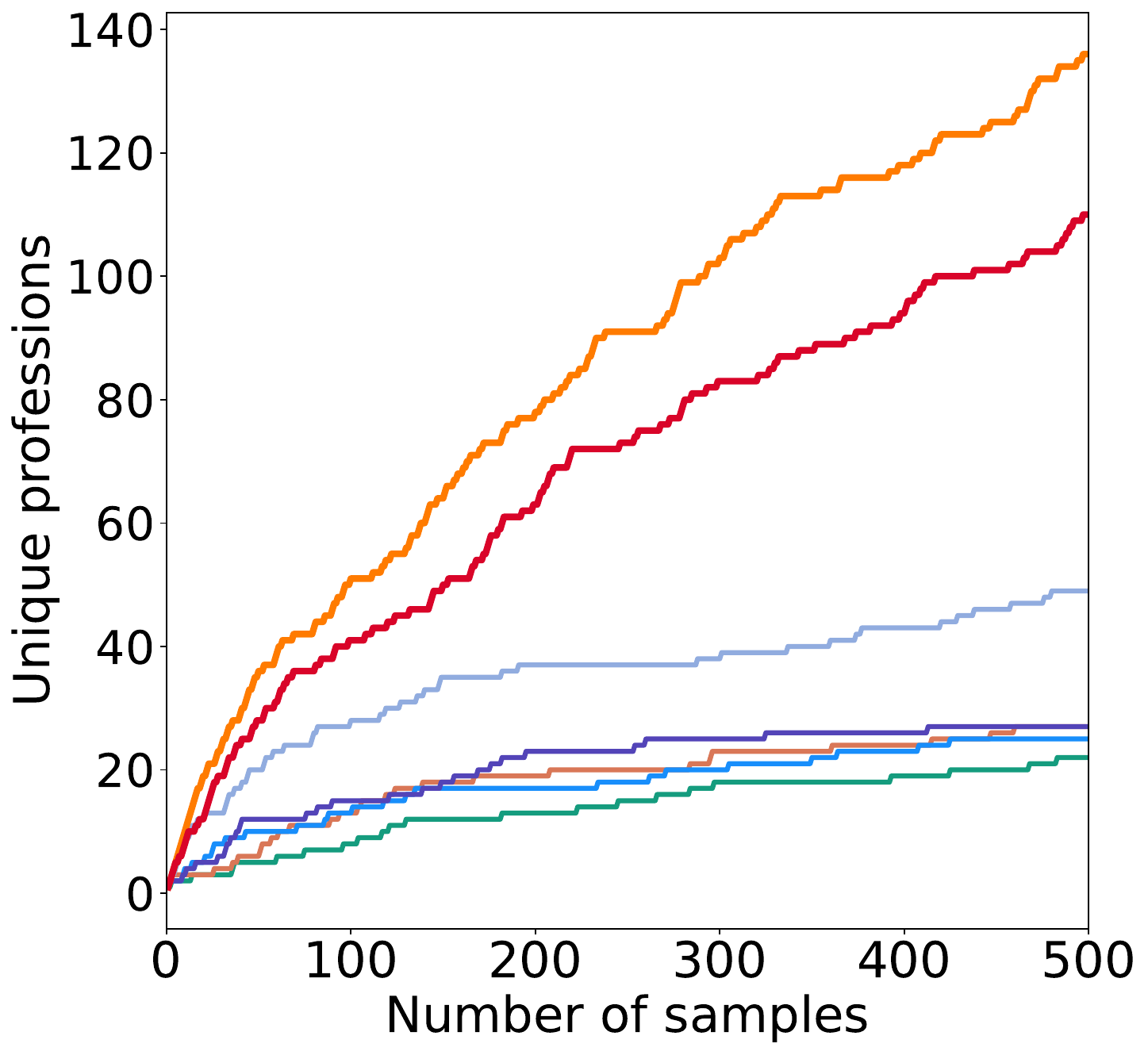}
    \caption{Professions.}
    \label{fig:fff}
  \end{subfigure}
  
  \begin{subfigure}[b]{0.32\textwidth}
    \includegraphics[width=\linewidth]{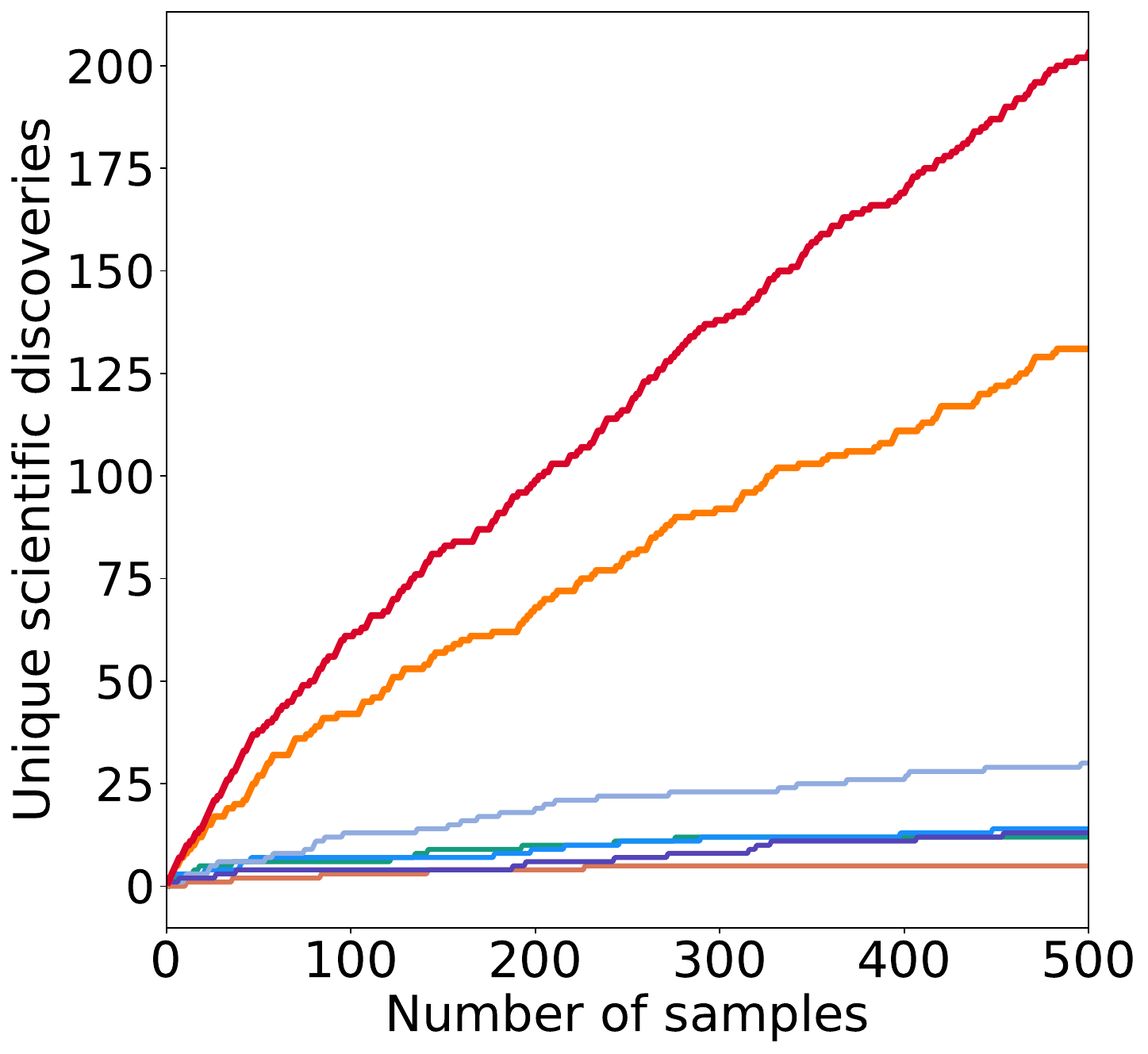}
    \caption{Scientific discoveries.}
    \label{fig:ggg}
  \end{subfigure}\hfill
  \begin{subfigure}[b]{0.32\textwidth}
    \includegraphics[width=\linewidth]{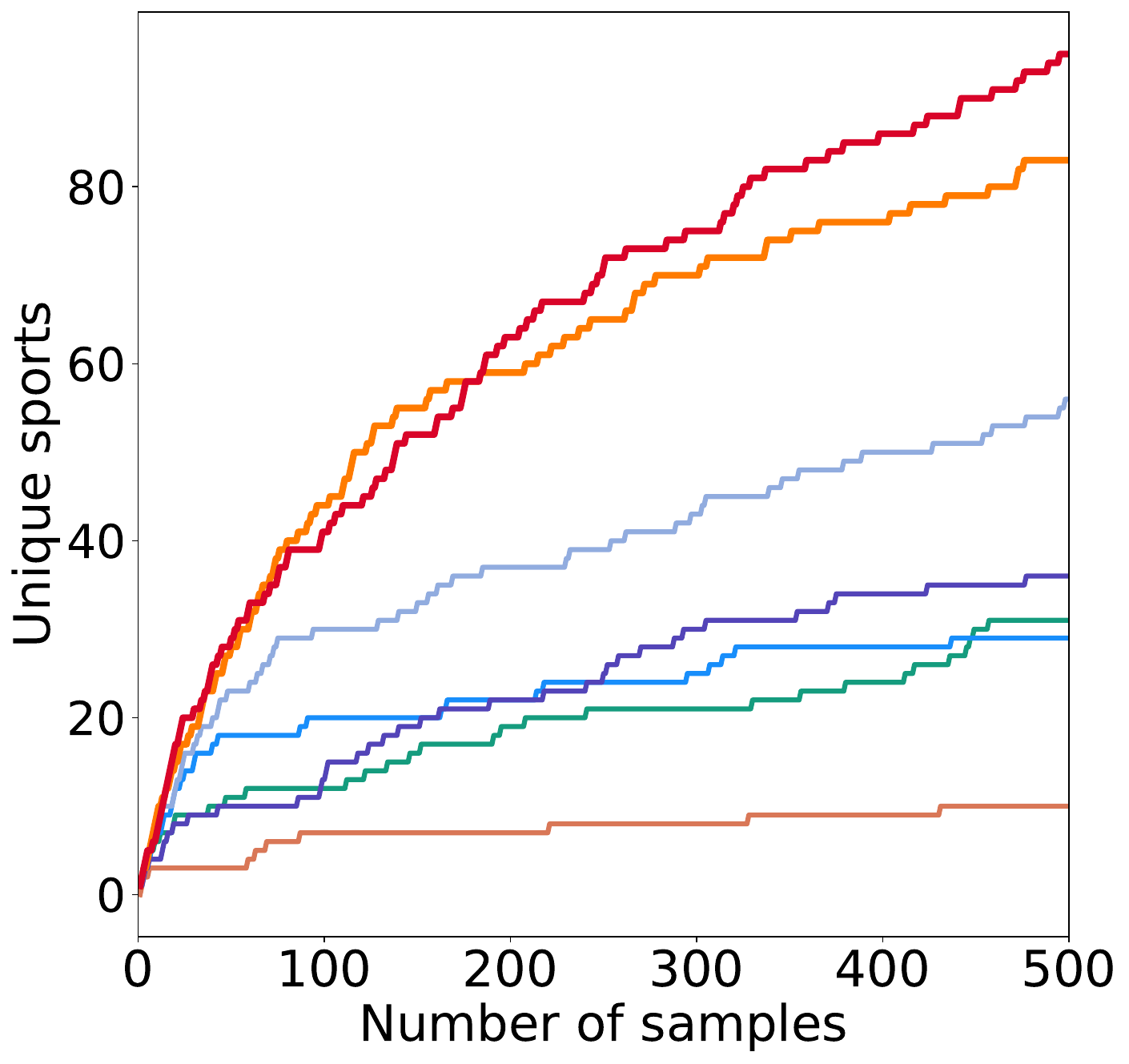}
    \caption{Sports.}
    \label{fig:rrr}
  \end{subfigure}\hfill
  \begin{subfigure}[b]{0.32\textwidth}
    \includegraphics[width=\linewidth]{images/openset/average_unique_samples_bar_chart.pdf}
    \caption{Avg. Unique@500.}
    \label{fig:plot3_appendix}
  \end{subfigure}
  
  \caption{\textbf{Open set diversity (Additional Results)}. (a)-(h) Cumulative unique responses when generating 500 samples for
open-ended prompts \eg, \textit{"Name a book title"}. (i) Average number of unique samples across categories.}
  \label{fig:open_set_app}
  \end{minipage}
  } 
\end{figure*}
\newpage
\section{Creativity - Diversity Visualization}
\label{app:responses_comp}
In this section, we show side-by-side comparisons of the responses generated by the Qwen2.5 32B Instruct model before and after LoRA finetune with our proposed GAPO.
Figures~\ref{tab:haiku},~\ref{tab:dialogue}~\ref{tab:jokes} and~\ref{tab:chorus} show the first ten generation per creative task prompt.
For each model the first ten responses are sorted alphabetically.
Reoccurring strings are shown in color to ease qualitative evaluation.
\begin{table*}[!htb]
\definecolor{cherrycolor}{RGB}{255,204,203}   
\definecolor{fallcolor}{RGB}{193,225,193}     
\definecolor{whispercolor}{RGB}{217,240,255}  
\definecolor{petalcolor}{RGB}{230,217,255}    
\definecolor{autumncolor}{RGB}{255,228,181}   
\definecolor{springcolor}{RGB}{255,255,224}   
\definecolor{silencecolor}{RGB}{255,219,255}  

\newcommand{\hlc}[2]{\colorbox{#1}{#2}}

\centering
\caption{Comparison of haikus generated by qwen2.5 32B baseline and GAPO models with the prompt: \textit{"Write a haiku in English."}.}
\begin{tabular}{p{0.44\textwidth}|p{0.44\textwidth}}
\toprule
\textbf{Qwen32B} & \textbf{Qwen32B GAPO} \\
\midrule

\hlc{cherrycolor}{Cherry blossoms} bloom, & \hlc{autumncolor}{Autumn} leaves \hlc{whispercolor}{whisper}, \\
Soft \hlc{whispercolor}{whispers} in the \hlc{springcolor}{spring} breeze, & Chill winds carry their secrets, \\
Beauty fades too soon. & \hlc{silencecolor}{Silent} paths remain. \\
\midrule

\hlc{cherrycolor}{Cherry blossoms} \hlc{fallcolor}{fall}, & \hlc{cherrycolor}{Cherry blossoms} bloom, \\
\hlc{whispercolor}{Whispering} \hlc{springcolor}{spring’s} ephemeral, & \hlc{petalcolor}{Petals} \hlc{fallcolor}{fall} like gentle snow, \\
Beauty fades, yet stays. & \hlc{silencecolor}{Spring} \hlc{whispercolor}{whispers} softly. \\
\midrule

\hlc{cherrycolor}{Cherry blossoms} \hlc{fallcolor}{fall}, & \hlc{cherrycolor}{Cherry blossoms} bloom, \\
\hlc{whispercolor}{Whispering} \hlc{springcolor}{spring’s} gentle goodbye, & Soft \hlc{whispercolor}{whispers} in the breeze sway, \\
\hlc{petalcolor}{Petals} carpet the earth. & Springtime’s fleeting dream. \\
\midrule

\hlc{cherrycolor}{Cherry blossoms} \hlc{fallcolor}{fall}, & \hlc{cherrycolor}{Cherry blossoms} \hlc{fallcolor}{fall}, \\
\hlc{whispercolor}{Whispering} \hlc{springcolor}{spring’s} soft farewell, & \hlc{whispercolor}{Whispers} of \hlc{springcolor}{spring} in the air, \\
\hlc{petalcolor}{Petals} drift like snow. & \hlc{petalcolor}{Petals} kiss the earth. \\
\midrule

\hlc{cherrycolor}{Cherry blossoms} \hlc{fallcolor}{fall}, & Leaves \hlc{whispercolor}{whisper} secrets, \\
\hlc{whispercolor}{Whispering} \hlc{springcolor}{spring’s} soft farewell, & \hlc{autumncolor}{Autumn}’s breath \hlc{whispercolor}{whispers} cold air, \\
\hlc{petalcolor}{Petals} drift like snow. & Dusk cloaks \hlc{silencecolor}{silent} world. \\
\midrule

\hlc{cherrycolor}{Cherry blossoms} \hlc{fallcolor}{fall}, & Leaves \hlc{whispercolor}{whisper} softly, \\
\hlc{whispercolor}{Whispering} \hlc{springcolor}{spring’s} soft goodbye, & \hlc{autumncolor}{Autumn}’s breath turns colors gold, \\
\hlc{petalcolor}{Petals} carpet the earth. & Dusk falls on the path. \\
\midrule

\hlc{cherrycolor}{Cherry blossoms} \hlc{fallcolor}{fall}, & Leaves \hlc{whispercolor}{whisper} softly, \\
\hlc{whispercolor}{Whispering} \hlc{springcolor}{spring’s} transient grace, & Chill \hlc{autumncolor}{Autumn} breeze sweeps in, \\
\hlc{petalcolor}{Petals} carpet the earth. & Day fades to twilight. \\
\midrule

\hlc{cherrycolor}{Cherry blossoms} \hlc{fallcolor}{fall}, & Moonlight bathes the trees, \\
\hlc{whispercolor}{Whispers} of \hlc{springcolor}{spring} fade to dust, & \hlc{silencecolor}{Silent} \hlc{whispercolor}{whispers} fill the night, \\
\hlc{silencecolor}{Silence} holds the breath. & Peace covers the earth. \\
\midrule

\hlc{cherrycolor}{Cherry blossoms} \hlc{fallcolor}{fall}, & Morning dew glistens, \\
\hlc{whispercolor}{Whispers} of \hlc{springcolor}{spring} linger on, & \hlc{whispercolor}{Whispers} of dawn through the leaves, \\
\hlc{silencecolor}{Silence} fills the air. & Silence cradles gold. \\
\midrule

\hlc{cherrycolor}{Cherry blossoms} \hlc{fallcolor}{fall}, & \hlc{whispercolor}{Whispering} wind chills, \\
\hlc{whispercolor}{Whispers} of \hlc{springcolor}{spring} scatter wide, & Leaves tumble with \hlc{silencecolor}{silent} grace, \\
\hlc{petalcolor}{Petals} touch the earth. & Night \hlc{whispercolor}{whispers} goodbye. \\

\bottomrule
\end{tabular}
\label{tab:haiku}
\end{table*}

\newpage
\begin{table*}[!htb]
  \centering
  \scalebox{0.92}{ 
  \begin{minipage}{\textwidth}
  \definecolor{reportcolor}{RGB}{255,204,203}      
  \definecolor{conclusioncolor}{RGB}{181,216,255}  
  \definecolor{graphscolor}{RGB}{214,233,255}      
  \definecolor{alicecolor}{RGB}{193,225,193}       
  \definecolor{bobcolor}{RGB}{255,230,204}         
  \definecolor{uniquecolor}{RGB}{230,217,255}      
  \definecolor{momsoncolor}{RGB}{255,219,255}      
  \definecolor{sarahtomcolor}{RGB}{230,217,255}    
  \definecolor{relationshipcolor}{RGB}{255,214,231} 
  \definecolor{eventcolor}{RGB}{217,240,255}       
  \definecolor{dailycolor}{RGB}{255,232,209}       
  \definecolor{springcolor}{RGB}{255,255,224}      

  \definecolor{festivalcolor}{RGB}{255,239,213}
  \definecolor{fireworkscolor}{RGB}{235,235,255}
  \definecolor{lockcolor}{RGB}{255,250,205}
  \definecolor{doorcolor}{RGB}{255,240,245}
  \definecolor{doublecolor}{RGB}{216,255,239}
  \definecolor{raincolor}{RGB}{224,255,255}
  \definecolor{umbrellacolor}{RGB}{230,245,255}
  \definecolor{officecolor}{RGB}{255,243,230}
  \definecolor{surprisecolor}{RGB}{255,230,243}
  \definecolor{partycolor}{RGB}{245,255,230}
  \definecolor{bettercolor}{RGB}{235,224,255}
  \definecolor{anniversarycolor}{RGB}{255,235,235}
  \definecolor{sunsetcolor}{RGB}{255,245,230}
  \definecolor{vegetablescolor}{RGB}{230,255,230}
  \definecolor{allcolor}{RGB}{230,230,255}
  \definecolor{incrediblecolor}{RGB}{220,230,255}
  \newcommand{\hlc}[2]{\colorbox{#1}{#2}}
  
  \caption{Comparison of dialogues generated by qwen2.5 32B baseline and GAPO models with the prompt:
\textit{“Compose a two-line dialogue between two characters.”}}
  \small
  \begin{tabular}{p{0.48\textwidth}|p{0.48\textwidth}}
    \toprule
    \textbf{Qwen32B} & \textbf{Qwen32B GAPO} \\
    \midrule

    \hlc{alicecolor}{Alice:} "\hlc{relationshipcolor}{Did you} \hlc{reportcolor}{finish the report?}" \newline
    \hlc{bobcolor}{Bob:} \hlc{graphscolor}{"Almost, just need to add the final graphs."}
    &
    " \hlc{relationshipcolor}{Did you} hear about the \hlc{festivalcolor}{festival} this weekend?" \newline
    "Yes, I'm excited to see the \hlc{fireworkscolor}{fireworks}!" \\
    \midrule

    \hlc{alicecolor}{Alice:} "\hlc{relationshipcolor}{Did you} \hlc{reportcolor}{finish the report?}" \newline
    \hlc{bobcolor}{Bob:} \hlc{graphscolor}{"Almost, just need to add the final graphs."}
    &
    "\hlc{relationshipcolor}{Did you} \hlc{uniquecolor}{lock the door} before we left?" \newline
    "Yeah, I double-checked it." \\
    \midrule

    \hlc{alicecolor}{Alice:} "\hlc{relationshipcolor}{Did you} \hlc{reportcolor}{finish the report?}" \newline
    \hlc{bobcolor}{Bob:} \hlc{conclusioncolor}{"Almost, just wrapping up the conclusions."}
    &
    "I can't believe it's \hlc{raincolor}{raining} again." \newline
    "Looks like we'll need our \hlc{umbrellacolor}{umbrellas} today." \\
    \midrule

    \hlc{alicecolor}{Alice:} "\hlc{relationshipcolor}{Did you} \hlc{reportcolor}{finish the report?}" \newline
    \hlc{bobcolor}{Bob:} \hlc{conclusioncolor}{"Almost, just wrapping up the conclusions."}
    &
    "Where have you been all night?" \newline
    "Sorry, I got caught up at the \hlc{officecolor}{office}." \\
    \midrule

    \hlc{alicecolor}{Alice:} "\hlc{relationshipcolor}{Did you} \hlc{reportcolor}{finish the report?}" \newline
    \hlc{bobcolor}{Bob:} \hlc{conclusioncolor}{"Almost, just wrapping up the conclusions."}
    &
    \hlc{alicecolor}{Alice:} "\hlc{relationshipcolor}{Did you} hear about the big \hlc{partycolor}{surprise party}?" \newline
    \hlc{bobcolor}{Bob:} "No way! Who's it for?" \\
    \midrule

    \hlc{alicecolor}{Alice:} "\hlc{relationshipcolor}{Did you} \hlc{reportcolor}{finish the report?}" \newline
    \hlc{bobcolor}{Bob:} \hlc{conclusioncolor}{"Almost, just wrapping up the conclusions."}
    &
    \hlc{alicecolor}{Alice:} "\hlc{relationshipcolor}{Did you} how was your day?" \newline
    \hlc{bobcolor}{Bob:} "It could've been better." \\
    \midrule

    \hlc{alicecolor}{Alice:} "\hlc{relationshipcolor}{Did you} \hlc{reportcolor}{finish the report?}" \newline
    \hlc{bobcolor}{Bob:} \hlc{conclusioncolor}{"Almost, just wrapping up the conclusions."}
    &
    \hlc{alicecolor}{Alice:} "\hlc{relationshipcolor}{Did you} forget our \hlc{anniversarycolor}{anniversary} again?"\newline
    \hlc{bobcolor}{Bob:} "Sorry, I promise to make it up to you this time." \\
    \midrule

    \hlc{alicecolor}{Alice:} "\hlc{relationshipcolor}{Did you} \hlc{reportcolor}{finish the report?}" \newline
    \hlc{bobcolor}{Bob:} \hlc{conclusioncolor}{"Almost, just wrapping up the conclusions."}
    &
    \hlc{alicecolor}{Alice:} "\hlc{relationshipcolor}{Did you} see the \hlc{sunsetcolor}{sunset} today?" \newline
    \hlc{bobcolor}{Bob:} "It always makes my day brighter, Alice." \\
    \midrule

    \hlc{alicecolor}{Alice:} "\hlc{relationshipcolor}{Did you} \hlc{uniquecolor}{lock the door}?" \newline
    \hlc{bobcolor}{Bob:} "Yes, I did, no need to worry."
    &
    Mom: "Have you eaten your \hlc{vegetablescolor}{vegetables}?" \newline
    Son: "Yes, Mom, all of them!" \\
    \midrule

    \hlc{alicecolor}{Alice:} "\hlc{relationshipcolor}{Did you} see the \hlc{surprisecolor}{eclipse} last night?" \newline
    \hlc{bobcolor}{Bob:} "No, I missed it. Was it spectacular?"
    &
    Sarah: "I can't believe we're finally here." \newline
    Tom: "It's incredible, isn't it?" \\
    \bottomrule
  \end{tabular}
  \label{tab:dialogue}
  \end{minipage}
  } 
\end{table*}
\newpage
\begin{table*}[!htb]
\definecolor{atomsetupcolor}{RGB}{255,204,203}    
\definecolor{alonaa}{RGB}{181,216,255}  
\definecolor{introcolor}{RGB}{193,225,193}     
\definecolor{computerjokecolor}{RGB}{255,230,204}  
\definecolor{tomatojokecolor}{RGB}{230,217,255}  
\definecolor{scarecrowjokecolor}{RGB}{255,219,255}  
\definecolor{mathjokecolor}{RGB}{217,240,255}    
\definecolor{salad}{RGB}{255,232,209}  
\newcommand{\hlc}[2]{\colorbox{#1}{#2}}
\caption{Comparison of jokes generated by qwen2.5 32B baseline and GAPO models with the prompt:
\textit{“Tell a joke.”}.}
\small
\begin{tabular}{p{0.48\textwidth}|p{0.48\textwidth}}
\toprule
\textbf{qwen32B} & \textbf{qwen32B GAPO} \\
\midrule
\hlc{introcolor}{Sure, here's a light joke for you:}\newline
\newline
\hlc{atomsetupcolor}{Why don't scientists trust atoms?}\newline
\newline
\hlc{atomsetupcolor}{Because they make up everything!} & 
\hlc{introcolor}{Sure! Here's a joke for you:}\newline
\newline
\hlc{salad}{Why did the computer go to the doctor?}\newline
\newline
\hlc{salad}{Because it had a virus and a byte to eat!} \\
\midrule

\hlc{atomsetupcolor}{Why don't scientists trust atoms?}\newline
\newline
\hlc{atomsetupcolor}{Because they make up everything.} & 
\hlc{introcolor}{Sure! Here's a joke for you:}\newline
\newline
\hlc{tomatojokecolor}{Why did the tomato turn red?}\newline
\newline
\hlc{tomatojokecolor}{Because it saw the salad dressing!}  \\
\midrule

\hlc{atomsetupcolor}{Why don't scientists trust atoms?}\newline
\newline
\hlc{atomsetupcolor}{Because they make up everything.} & 
\hlc{introcolor}{Sure! Here's a joke for you:}\newline
\newline
\hlc{tomatojokecolor}{Why did the tomato turn red?}\newline
\newline
\hlc{tomatojokecolor}{Because it saw the salad dressing!}  \\
\midrule

\hlc{atomsetupcolor}{Why don't scientists trust atoms?}\newline
\newline
\hlc{atomsetupcolor}{Because they make up everything.} & 
\hlc{introcolor}{Sure! Here's a joke for you:}\newline
\newline
\hlc{atomsetupcolor}{Why don't scientists trust atoms?}\newline
\newline
\hlc{atomsetupcolor}{Because they make up everything!} \\
\midrule

\hlc{atomsetupcolor}{Why don't scientists trust atoms?}\newline
\newline
\hlc{atomsetupcolor}{Because they make up everything.} & 
\hlc{introcolor}{Sure! Here's a joke for you:}\newline
\newline
\hlc{atomsetupcolor}{Why don't scientists trust atoms?}\newline
\newline
\hlc{atomsetupcolor}{Because they make up everything.}  \\
\midrule

\hlc{atomsetupcolor}{Why don't scientists trust atoms?}\newline
\newline
\hlc{atomsetupcolor}{Because they make up everything.} & 
\hlc{introcolor}{Sure! Here's one for you:}\newline
\newline
\hlc{alonaa}{Why did the scarecrow win an award?}\newline
\newline
\hlc{alonaa}{Because he was outstanding in his field!}  \\
\midrule

\hlc{atomsetupcolor}{Why don't scientists trust atoms?}\newline
\newline
\hlc{atomsetupcolor}{Because they make up everything.} & 
\hlc{atomsetupcolor}{Why don't scientists trust atoms?}\newline
\newline
\hlc{atomsetupcolor}{Because they make up everything!} \\
\midrule

\hlc{atomsetupcolor}{Why don't scientists trust atoms?}\newline
\newline
\hlc{atomsetupcolor}{Because they make up everything.} & 
\hlc{atomsetupcolor}{Why don't scientists trust atoms?} \hlc{atomsetupcolor}{Because they make up everything!} \\
\midrule

\hlc{atomsetupcolor}{Why don't scientists trust atoms?}\newline
\newline
\hlc{atomsetupcolor}{Because they make up everything.} & 
\hlc{atomsetupcolor}{Why don't scientists trust atoms?} \hlc{atomsetupcolor}{Because they make up everything!}  \\
\midrule

\hlc{atomsetupcolor}{Why don't scientists trust atoms?}\newline
\newline
\hlc{atomsetupcolor}{Because they make up everything.} & 
\hlc{scarecrowjokecolor}{Why was the math book unhappy?}  
\newline
\newline
\hlc{scarecrowjokecolor}{Because it had too many problems!}  \\
\bottomrule
\end{tabular}
\label{tab:jokes}
\end{table*}
\begin{table*}
\definecolor{shinebrightcolor}{RGB}{255,204,203}  
\definecolor{moonlightcolor}{RGB}{181,216,255}    
\definecolor{feelsrightcolor}{RGB}{193,225,193}   
\definecolor{dancingcolor}{RGB}{255,230,204}      
\definecolor{ohcolor}{RGB}{230,217,255}           
\definecolor{lovecolor}{RGB}{255,219,255}         
\definecolor{ipronouncolor}{RGB}{217,240,255}     
\definecolor{babycolor}{RGB}{255,255,224}         

\newcommand{\hlc}[2]{\colorbox{#1}{#2}}
\caption{Comparison of song choruses generated by qwen2.5 32B baseline and GAPO models with the prompt:
\textit{“Write only the chorus for a pop song.”}.}
\small
\begin{tabular}{p{0.48\textwidth}|p{0.48\textwidth}}
\toprule
\textbf{Qwen32B} & \textbf{Qwen32B GAPO} \\
\midrule
Chorus:\newline
\hlc{shinebrightcolor}{Shine bright} like a diamond in the night,\newline
\hlc{dancingcolor}{Dancing} under the \hlc{moonlightcolor}{moonlight}, everything \hlc{feelsrightcolor}{feels right},\newline
Jump into the rhythm, let your soul take flight,\newline
\hlc{shinebrightcolor}{Shine bright}, oh so bright, make the darkness take a dive tonight. & 
All \hlc{ipronouncolor}{I} need is a little bit more,\newline
In this moment, you're my only door,\newline
\hlc{ohcolor}{Oh}, let's \hlc{dancingcolor}{dance} in the \hlc{moonlightcolor}{moonlight} glow,\newline
Together, we'll make it all worthwhile, yeah! \\
\midrule

\hlc{dancingcolor}{Dancing} in the \hlc{moonlightcolor}{moonlight}, feeling so alive,\newline
Every moment's right, under this endless sky.\newline
Heartbeats sync with the stars, \hlc{shinebrightcolor}{shining bright} tonight,\newline
Lost in the rhythm, everything \hlc{feelsrightcolor}{feels alright}. & 
\hlc{ipronouncolor}{I'll} hold on forever, never letting go\newline
Our \hlc{lovecolor}{love} is like a fire, burning so bright\newline
We'll \hlc{dancingcolor}{dance} through all the night and find our way\newline
\hlc{lovecolor}{Love}'s the only thing that can make us stay \\
\midrule

Here's the sun, \hlc{shinebrightcolor}{shining} down on me,\newline  
In this moment, \hlc{ipronouncolor}{I} feel completely free.\newline  
Every star, a whisper in the dark,\newline  
Together we spark. & 
\hlc{ipronouncolor}{I'm} alive, \hlc{ipronouncolor}{I} believe in me, \hlc{shinebrightcolor}{shining bright} and free,\newline
Every moment \hlc{feelsrightcolor}{feels so right} under this endless sky, yeah! \\
\midrule

Here's the sun, \hlc{shinebrightcolor}{shining} through,\newline 
Got that feeling, brand new,\newline
\hlc{dancingcolor}{Dancing} under the moon,\newline 
This moment's forever, or so it seems. & 
\hlc{ipronouncolor}{I'm} falling in \hlc{lovecolor}{love} all over again\newline
Won't let go of this feeling once it starts\newline
\hlc{ipronouncolor}{I'm} falling in \hlc{lovecolor}{love} all over again\newline
Can't deny this moment, it's right where we are \\
\midrule

Let's \hlc{dancingcolor}{dance} under the \hlc{moonlightcolor}{moonlight}, feel the rhythm of the night,\newline
Together we \hlc{shinebrightcolor}{shine so bright}, everything \hlc{feelsrightcolor}{feels right}. & 
\hlc{ipronouncolor}{I'm} just gonna \hlc{dancingcolor}{dance} with you tonight\newline
Feel the rhythm, let it ignite\newline
In this moment, we're alive\newline
\hlc{dancingcolor}{Dancing} into the night, oh my life \\
\midrule

\hlc{ohcolor}{Oh oh oh}, let the music move your soul tonight,\newline
Under the \hlc{moonlightcolor}{moonlight}, everything feels brand new,\newline
\hlc{ohcolor}{Oh oh oh}, \hlc{dancingcolor}{dance} like nobody's watching you,\newline
In this moment, we're alive, \hlc{shinebrightcolor}{shining so bright}. & 
Let's \hlc{dancingcolor}{dance} tonight\newline
No more hiding in the light\newline
It's time to feel alive\newline
\hlc{shinebrightcolor}{Shine bright} in the night sky with you and I \\
\midrule

\hlc{ohcolor}{Oh oh}, we \hlc{shinebrightcolor}{shine so bright} tonight,\newline
Under the \hlc{moonlightcolor}{moonlight}, everything \hlc{feelsrightcolor}{feels right},\newline
Hand in hand, we \hlc{dancingcolor}{dance} into the light,\newline
\hlc{ohcolor}{Oh oh}, everything's gonna be alright. & 
Let's go \hlc{dancingcolor}{dancing} in the \hlc{moonlightcolor}{moonlight}\newline
Our hearts beating as one tonight\newline
Feel the rhythm and let it shine\newline
\hlc{dancingcolor}{Dancing} in the \hlc{moonlightcolor}{moonlight}, we're just fine \\
\midrule

\hlc{ohcolor}{Oh oh}, we \hlc{shinebrightcolor}{shine so bright} tonight,\newline
Underneath this endless sky, everything \hlc{feelsrightcolor}{feels right},\newline
Hand in hand, we \hlc{dancingcolor}{dance} into the light,\newline
\hlc{ohcolor}{Oh oh}, together we ignite. & 
\hlc{ohcolor}{Oh} \hlc{babycolor}{baby}, let's \hlc{dancingcolor}{dance} all night long\newline
In this moment we're right where we belong\newline
Under the \hlc{moonlightcolor}{moonlight}, feeling so right\newline
Let's keep \hlc{dancingcolor}{dancing} until the morning light comes along \\
\midrule

\hlc{ohcolor}{Oh}, let the music move your body tonight,\newline
\hlc{dancingcolor}{Dancing} under the stars, everything \hlc{feelsrightcolor}{feels right},\newline
Heartbeats in rhythm with the \hlc{moonlightcolor}{moonlight},\newline
\hlc{shinebrightcolor}{Shine on}, \hlc{shinebrightcolor}{shine on}, till the morning light. & 
\hlc{ohcolor}{Oh} \hlc{babycolor}{baby}, we're \hlc{dancingcolor}{dancing} under the stars tonight\newline
We won't ever let this moment go\newline
Our \hlc{lovecolor}{love} is \hlc{shinebrightcolor}{shining brighter} than the light\newline
Come closer, feel the magic grow, oh yeah \\
\midrule

We \hlc{shinebrightcolor}{shine so bright}, night turns to day,\newline 
In this moment, let's lose all sense of way,\newline 
Hearts beating loud, under the starlit sky,\newline 
Together we fly, where the dreamers lie. & 
\hlc{ohcolor}{Oh} \hlc{babycolor}{baby}, you light up my world so bright\newline
Every moment feels like a never-ending night\newline
\hlc{ipronouncolor}{I} just wanna hold on tight\newline
\hlc{ohcolor}{Oh} \hlc{babycolor}{baby}, you light up my world, \hlc{shinebrightcolor}{shining right} tonight \\
\bottomrule
\end{tabular}
\label{tab:chorus}
\end{table*}
\newpage
\newpage
\newpage
\section{Creativity and Coherence}
\label{app:creativity_coherence}

\begin{figure}[htbp]
    \centering

    \begin{subfigure}[b]{0.45\textwidth}
        \includegraphics[width=\linewidth]{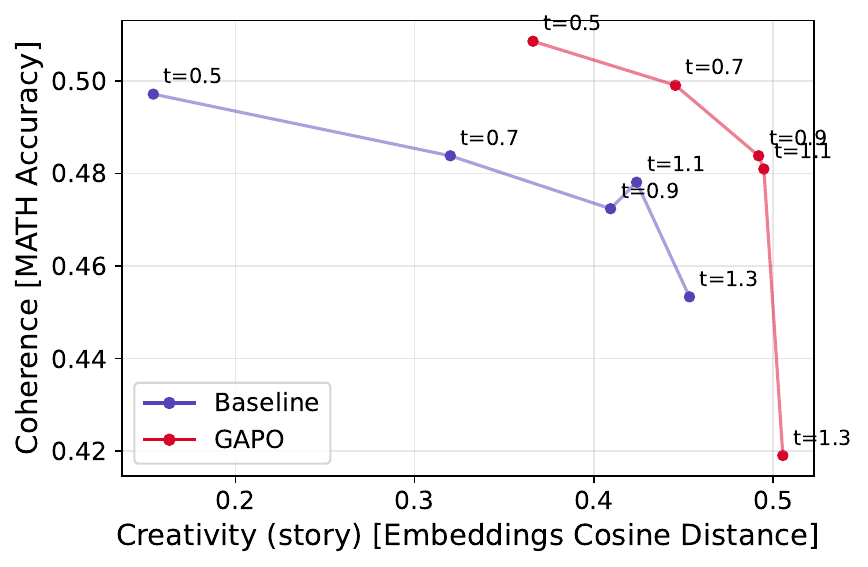}
        \caption{Story.}
    \end{subfigure}
    \hfill
    \begin{subfigure}[b]{0.45\textwidth}
        \includegraphics[width=\linewidth]{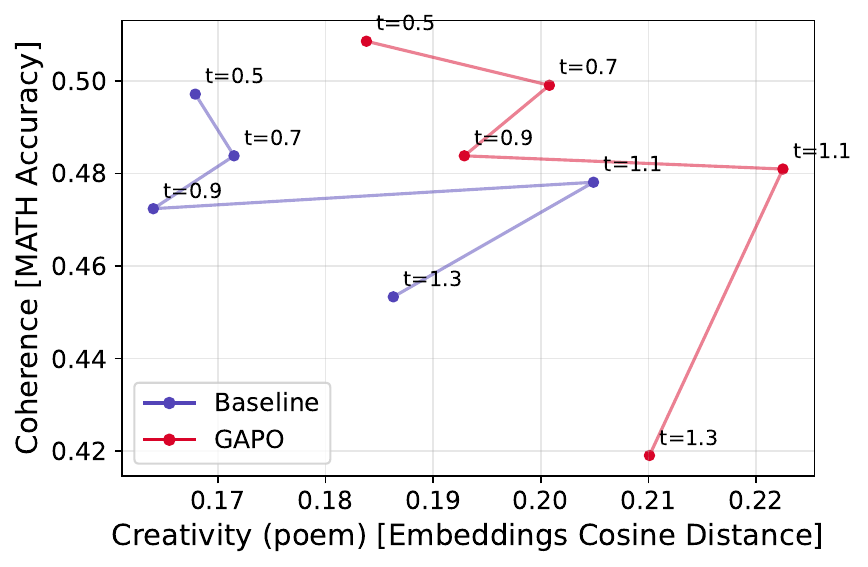}
        \caption{Poem.}
    \end{subfigure}

    \vspace{0.5cm}

    \begin{subfigure}[b]{0.45\textwidth}
        \includegraphics[width=\linewidth]{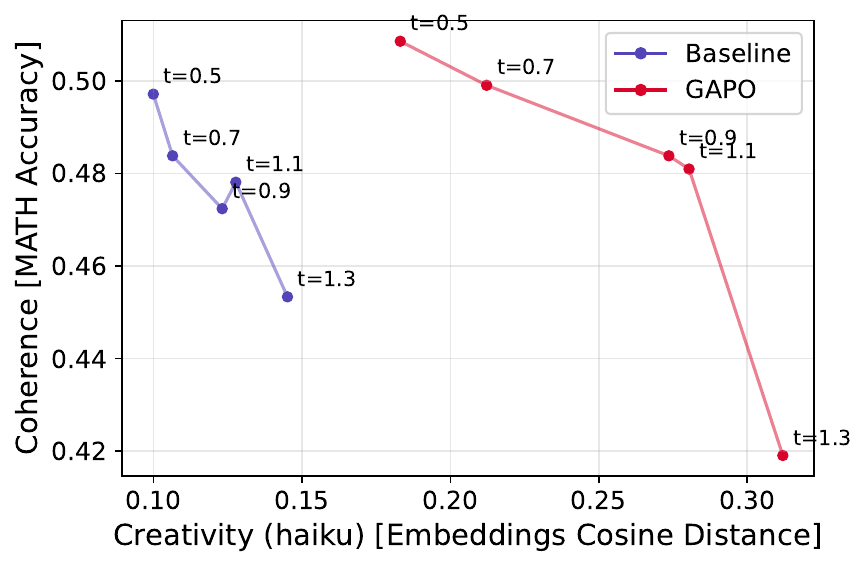}
        \caption{Haiku.}
    \end{subfigure}
    \hfill
    \begin{subfigure}[b]{0.45\textwidth}
        \includegraphics[width=\linewidth]{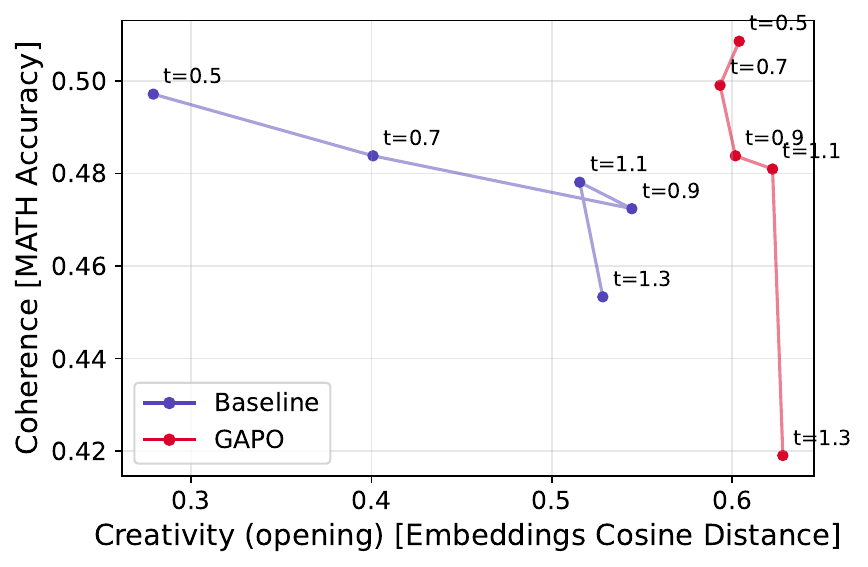}
        \caption{Opening.}
    \end{subfigure}

    \vspace{0.5cm}

    \begin{subfigure}[b]{0.45\textwidth}
        \includegraphics[width=\linewidth]{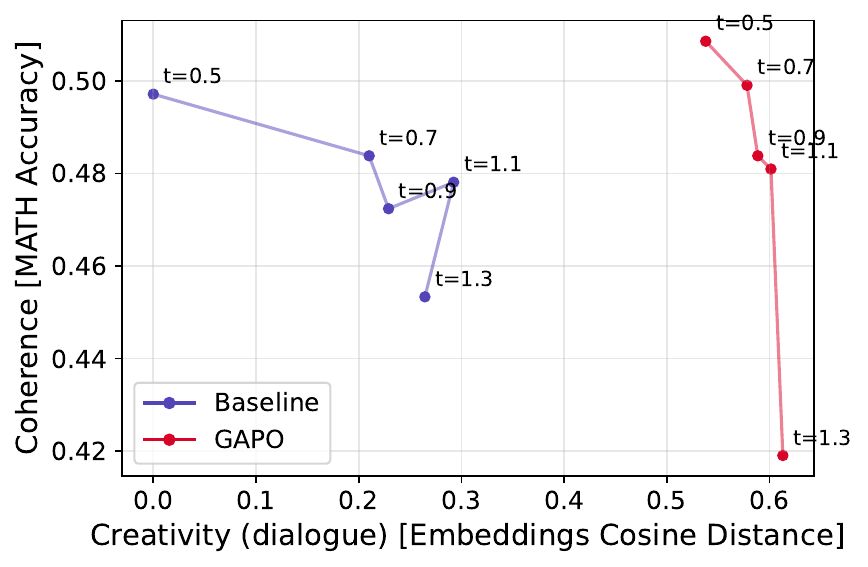}
        \caption{Dialogue.}
    \end{subfigure}
    \hfill
    \begin{subfigure}[b]{0.45\textwidth}
        \includegraphics[width=\linewidth]{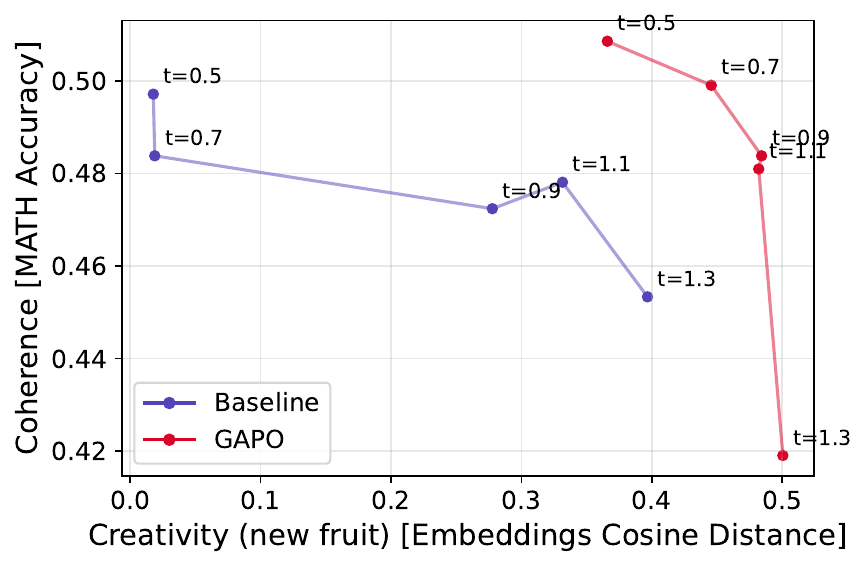}
        \caption{New Fruit.}
    \end{subfigure}

    \vspace{0.5cm}

    \begin{subfigure}[b]{0.45\textwidth}
        \includegraphics[width=\linewidth]{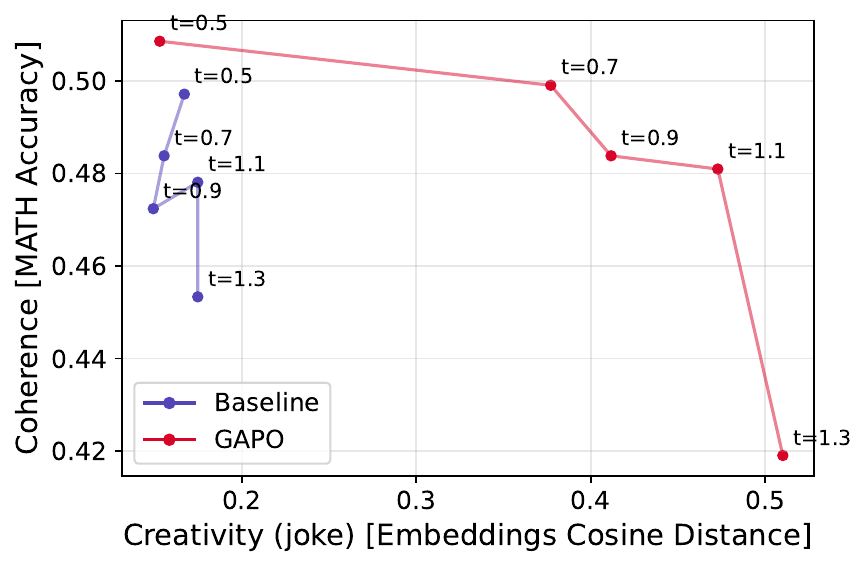}
        \caption{Joke.}
    \end{subfigure}
    \hfill
    \begin{subfigure}[b]{0.45\textwidth}
        \includegraphics[width=\linewidth]{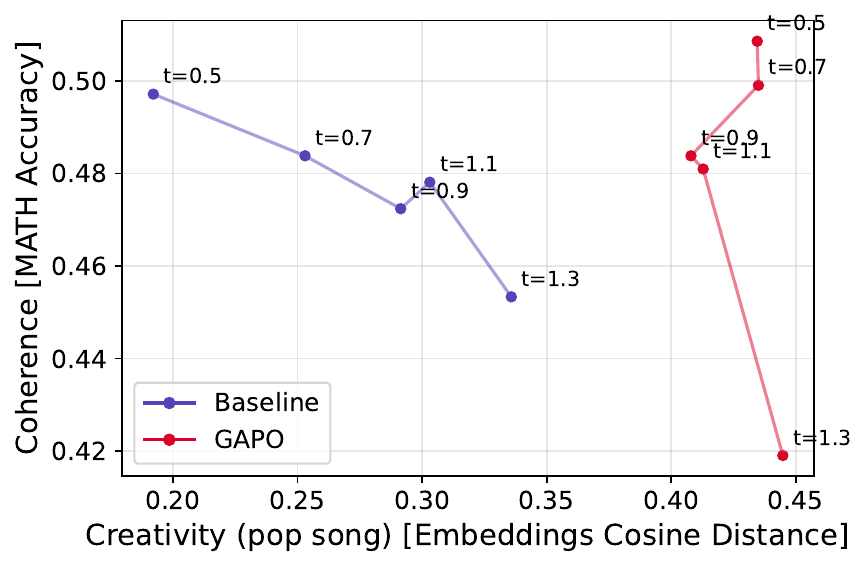}
        \caption{Pop Song.}
    \end{subfigure}

    \caption{\textbf{Coherence vs. Creativity additional results.}}
    \label{fig:fig_grid}
\end{figure}

We provide additional results in Figure~\ref{fig:fig_grid} to illustrate the effects of GAPO on creativity and coherence. Each plot shows results for a single creative prompt from Table~\ref{tab:text_creativity}, while the results in Section~\ref{sec:creativity_coherence_tradeoff} present the average creativity metric across prompts.
Compared to the baseline, GAPO generates more varied and creative responses.

\newpage
\section{Training Data Examples}
\label{app:app_training_examples}
In the following table, we present a collection of training data examples organized by prompt and topic. Each prompt instructs the selection of a single item from a list, with examples drawn from various categories. The complete set of categories in the dataset includes: Animals, Countries, Emotions, English, Foods, Letters, Numbers, and Vehicles. All examples follow a consistent pattern where the model is expected to make a selection and format its response using the specified XML tags. The examples shown below represent a subset from three of these categories.
\begin{table*}[!hb]
\centering
\scalebox{0.8}{  
\begin{tabular}{p{0.85\textwidth}|p{0.1\textwidth}}
\toprule
\textbf{Prompt} & \textbf{Topic} \\
\midrule
``Pick exactly one option, ensuring uniform selection: [White Ibis, Hawk Moth, Coelacanth, Crossbill, Grebe, Guinea Fowl]. Format your response as follows: $<$answer$>$selected\_item$<$/answer$>$.'' & Animals \\
\midrule
``Instruction: Please pick one from the following list: [Softshell Turtle, Oriental Stork, Harpy Eagle, Barracuda, Nighthawk, Sparrowhawk, Manatee, Planarian, Squacco Heron, Bittern, Snipe]. Format your response as follows: $<$answer$>$selected\_item$<$/answer$>$.'' & Animals \\
\midrule
``Please select one from these options: [Grizzly Bear, Hermit Crab, Spotted Owl, Jerboa, Honeybee]. Format your response as follows: $<$answer$>$selected\_item$<$/answer$>$.'' & Animals \\
\midrule
``Take one from these choices: [Hawksbill Turtle, Wildebeest, Gaur, Thick-knee, Cormorant]. Format your response as follows: $<$answer$>$selected\_item$<$/answer$>$.'' & Animals \\
\midrule
``Pick exactly one option, ensuring uniform selection: [Reverent, Calm, Sympathetic, Intrigued, Astonished, Disgusted]. Format your response as follows: $<$answer$>$selected\_item$<$/answer$>$.'' & Emotions \\
\midrule
``Please randomly select one option with equal probability: [Disappointed, Delighted, Rejected, Amused, Disgusted, Triumphant, Captivated]. Format your response as follows: $<$answer$>$selected\_item$<$/answer$>$.'' & Emotions \\
\midrule
``Pick one option from the list: [Envious, Perplexed, Tense, Nostalgic, Impatient]. Format your response as follows: $<$answer$>$selected\_item$<$/answer$>$.'' & Emotions \\
\midrule
``Please choose one option from the list: [Ecstatic, Uncomfortable, Disappointed, Proud, Overwhelmed, Surprised, Perplexed, Fascinated, Intimidated, Apathetic]. Format your response as follows: $<$answer$>$selected\_item$<$/answer$>$.'' & Emotions \\
\midrule
``Pick exactly one option, ensuring uniform selection: [60, 8362, 8990, 4265, 7731, 2817]. Format your response as follows: $<$answer$>$selected\_item$<$/answer$>$.'' & Numbers \\
\midrule
``Instruction: Select one from these choices: [8330, 2258, 6507, 7349, 1908, 6383, 285, 6115, 9238]. Format your response as follows: $<$answer$>$selected\_item$<$/answer$>$.'' & Numbers \\
\midrule
``Select exactly one option at random: [7513, 6115, 7899, 5540, 115, 4733, 4262, 4425, 5778]. Format your response as follows: $<$answer$>$selected\_item$<$/answer$>$.'' & Numbers \\
\midrule
``Please choose just one from the list [7634, 5133, 6974, 7736]. Format your response as follows: $<$answer$>$selected\_item$<$/answer$>$.'' & Numbers \\
\bottomrule
\end{tabular}
}
\caption{Training data examples by prompt and topic.}
\label{tab:training_data}
\end{table*}

\newpage
\section{Implementation Details}
\label{app:app_implementation_details}

\paragraph{Framework and Architecture}
We implemented GAPO as a modification of the original GRPO method \cite{shao2024deepseekmath} using the HuggingFace Transformer Reinforcement Learning (TRL) framework \cite{vonwerra2022trl}.

\paragraph{Model Selection and Training}
In our experiments, we utilized the 7B and 32B Instruct variants of the Qwen2.5 family \cite{yang2024qwen2}. Each model was fine-tuned with GAPO on the dataset described in Appendix~\ref{app:app_training_examples}, with batch size 8 and learning rate 1e-5. For the training process, we employed Low-Rank Adaptation (LoRA) \cite{hu2022lora} with rank 64, alpha 32, and dropout 0.1. For GAPO we utilized 32 generations per group and no KL penalty on divergence from the reference policy ($\beta=0$).

\paragraph{Benchmark Evaluations}
We utilized the Language Model Evaluation Harness framework \cite{eval-harness} to conduct the benchmark evaluations. Specifically, the test subsets of the benchmarks we have utilized were sampled as follows:
\begin{itemize}
    \item GSM8K: 200 randomly sampled problems
    \item MATH: 210 samples (30 samples per each of the 7 sub-tasks)
    \item MMLU-Pro: 196 samples (14 samples per each of the 14 sub-tasks)
    \item HumanEval: Full evaluation set
\end{itemize}

Finally, we note that our code will be made publicly available to facilitate reproducibility and further research in this area.

\newpage
\section{Supervised Fine Tuning Baseline}
An alternative approach to address sampling bias in LLMs is to incorporate a teacher-forcing objective into the supervised fine-tuning (SFT) process.
Specifically, for each prompt, we construct all valid completions by appending each item from a reference list to the prompt. We then compute the next-token prediction loss for each completion and aggregate these losses. Minimizing the total loss encourages the model to assign similar probabilities to multiple valid outputs, thus promoting output diversity.

We experimented with the above baseline and compared its output diversity and coherence to those of the reward-based model, as shown in Table~\ref{tab:model_comparison}.
As observed, while the SFT baseline significantly improves diversity on in-distribution data compared to a vanilla Qwen2.5 model (\eg, reducing the Jensen-Shannon divergence from 0.31 to 0.19), it fails to generalize to unseen lists and tasks, as reflected by the Unique@500 metric, computing how many unique samples exist across 500 generated samples. This observation aligns with the findings reported in~\cite{chu2025sftmemorizesrlgeneralizes}.

\begin{table*}[h]  
  \centering
  \caption{Comparison of uniformity metrics between different models.}
  \label{tab:model_comparison}
  \renewcommand{\arraystretch}{1.5}
  \small
  \begin{tabular}{l|c|c}
    \toprule
    \textbf{Model}       & \textbf{JS} ($\downarrow$) & \textbf{Unique@500} ($\uparrow$) \\
    \midrule
    Qwen2.5 7B               &    0.31     &        29      \\
    +Min-p (0.05)            &    0.33     &        10      \\
    +Min-p (0.1)             &    0.36     &        6       \\
    +SFT                     &    0.19     &        3       \\
    +GAPO                    &\textbf{0.09}&  \textbf{112}  \\       
    \bottomrule
  \end{tabular}
\end{table*}
\newpage
\section{Licensing and Additional Disclosures}
\label{appendix:licensing}

\subsection{Artifact Licensing}
\label{appendix:artifact-licensing}

\noindent\textbf{Models.} 
Qwen2.5 7B and 32B Instruct models are licensed under Apache 2.0, permitting research use and modification.

\noindent\textbf{Frameworks.} 
HuggingFace TRL (Apache 2.0), LoRA/PEFT (Apache 2.0), Language Model Evaluation Harness (MIT).

\noindent\textbf{Datasets.} 
GSM8K (MIT License), MATH, HumanEval, and MMLU-Pro (academic research use). All usage complies with respective license terms.

\subsection{Synthetic Data Compliance}
\label{appendix:data-compliance}

Our training dataset consists entirely of synthetically generated lists from neutral categories (animals, countries, emotions, numbers, vehicles, foods, letters, English words). No personally identifiable information, copyrighted content, or real user data was incorporated. List items contain only factual, publicly available information.

\subsection{Code and Data Availability}
\label{appendix:code-availability}

Complete implementation including GAPO modifications to GRPO, training scripts, evaluation protocols, and synthetic dataset generation will be released under an open-source license to ensure reproducibility.

\end{document}